\documentclass{article}

\usepackage{microtype}
\usepackage{graphicx}
\usepackage[normalem]{ulem}
\useunder{\uline}{\ul}{}
\usepackage{booktabs} % for professional tables

\usepackage{hyperref}

\newcommand{\myurl}[1]{\href{#1}{URL}} % iñaki
\newcommand{\dline}{%
  \noalign{\vspace{3pt}}%
  \cdashline{1-6}[0.5pt/2pt]%
  \noalign{\vspace{3pt}}%
}

\usepackage[preprint]{icml2026}

\usepackage{amsmath}
\usepackage{amssymb}
\usepackage{mathtools}
\usepackage{amsthm}

\usepackage[capitalize,noabbrev]{cleveref}

\theoremstyle{plain}

\theoremstyle{definition}

\theoremstyle{remark}

\usepackage[textsize=tiny]{todonotes}
\usepackage{longtable}
\usepackage{booktabs}
\usepackage{url}
\usepackage{subcaption}   % loads caption
\usepackage{caption}
\usepackage{array}
\usepackage{arydshln}
\usepackage{booktabs,longtable,array,ragged2e,url}
\usepackage[T1]{fontenc} 
\usepackage[utf8]{inputenc}
\newcolumntype{L}[1]{>{\raggedright\let\newline\\\arraybackslash\hspace{0pt}}p{#1}}

\icmltitlerunning{MrBERT: Modern Multilingual Encoders via Vocabulary, Domain, and
Dimensional Adaptation}

\begin{document}

\twocolumn[
\icmltitle{MrBERT: Modern Multilingual Encoders via Vocabulary, Domain, and Dimensional Adaptation}
\icmlsetsymbol{equal}{*}
\begin{icmlauthorlist}
\icmlauthor{Daniel Tamayo$^{*1}$}{}
\icmlauthor{Iñaki Lacunza$^{*1}$}{}
\icmlauthor{Paula Rivera-Hidalgo$^{*1}$}{} \\
\icmlauthor{Severino Da Dalt$^{1}$}{}
\icmlauthor{Javier Aula-Blasco$^{1}$}{}
\icmlauthor{Aitor Gonzalez-Agirre$^{1}$}{}
\icmlauthor{Marta Villegas$^{1}$}{}
\end{icmlauthorlist}

\begin{center}
\small
$^{1}$Barcelona Supercomputing Center
\end{center}

\icmlcorrespondingauthor{Aitor Gonzalez-Agirre}{langtech@bsc.es}
\icmlkeywords{Machine Learning, ICML}
\vskip 0.3in
]

% Suppress the default bottom-left affiliation/notice block
% \renewcommand{\printAffiliationsAndNotice}[1]{}
\printAffiliationsAndNotice{$^{*}$Core contribution}

% Add the core contribution line as a page footnote
% \twocolumn[
%   \icmltitle{MrBERT: Modern Multilingual Encoders via Vocabulary, Domain, and Dimensional Adaptation}

%   \begin{icmlauthorlist}
%     \icmlauthor{Daniel Tamayo$^{1*}$}{}
%     \icmlauthor{Iñaki Lacunza$^{1*}$}{}
%     \icmlauthor{Paula Rivera-Hidalgo$^{1*}$}{} \\
%     \icmlauthor{Severino Da Dalt$^{1}$}{}
%     \icmlauthor{Javier Aula-Blasco$^{1}$}{}
%     \icmlauthor{Aitor Gonzalez-Agirre$^{1}$}{}
%     \icmlauthor{Marta Villegas$^{1}$}{}
%   \end{icmlauthorlist}

%   \begin{center}
%   \small
%   $^1$Barcelona Supercomputing Center (BSC), Barcelona, Spain\\
%   \vspace{2pt}
%   {\footnotesize $^\dag$Work done while at BSC. Machine Learning and Optimization Laboratory, EPFL, Lausanne, Switzerland.}
%   \end{center}

%   \icmlcorrespondingauthor{Aitor Gonzalez-Agirre}{langtech@bsc.es}
%   \icmlkeywords{Machine Learning, ICML}

%   \vskip 0.3in
% ]

% \renewcommand{\printAffiliationsAndNotice}[1]{
%   \footnotetext{* Core Contributors. Correspondence: \texttt{langtech@bsc.es}}
% }
% \printAffiliationsAndNotice{}

% \printAffiliationsAndNotice{} % no special notice (required even if empty)
% Or, if applicable, use the standard equal contribution text:
% \printAffiliationsAndNotice{\icmlEqualContribution}

\begin{abstract}
  We introduce MrBERT, a family of 150M-300M parameter encoders built on the ModernBERT architecture and pre-trained on 35 languages and code. Through targeted adaptation, this model family achieves state-of-the-art results on Catalan- and Spanish-specific tasks, while establishing robust performance across specialized biomedical and legal domains. To bridge the gap between research and production, we incorporate Matryoshka Representation Learning (MRL), enabling flexible vector sizing that significantly reduces inference and storage costs. Ultimately, the MrBERT family demonstrates that modern encoder architectures can be optimized for both localized linguistic excellence and efficient, high-stakes domain specialization. We open source the complete model family on \href{https://huggingface.co/collections/BSC-LT/mrbert}{HuggingFace}.
\end{abstract}

\section{Introduction}

The transformer encoder architecture, initiated by BERT~\cite{BERT}, remains the standard for modern natural language understanding (NLU), serving as the foundation for successful models like RoBERTa~\cite{liu2019roberta} and XLM-RoBERTa~\cite{xlm-roberta}. While the prevailing research trend has shifted toward massive decoder-only models, recent advancements have successfully extended encoder capabilities to long-context and retrieval-heavy regimes. These developments, seen in models such as ModernBERT~\cite{modernbert}, mmBERT~\cite{mmbert} and mGTE~\cite{mgte}, deliver the high-quality representations required for large-scale inference without the efficiency trade-offs of generative frameworks.

% Despite these developments, a significant challenge persists in reconciling broad multilingual coverage with the rigorous requirements of high-stakes specialization. While specialized encoders have been developed for the biomedical \cite{ClinicalModernBERT, bioclinicalModernBERT} and legal \cite{legal-bert, PT_modernbert} domains, these models often remain decoupled from the architectural improvements seen in modern general-purpose encoders. We argue that the optimal path to specialization depends on the target context. For regional languages such as Spanish and Catalan, efficiency is best achieved through vocabulary adaptation \cite{flor} and language-specific data mining \cite{RobertaCA, roberta-base-bne, rigoberta}, allowing for more compact, task-optimized footprints. In contrast, for high-entropy domains like law, preserving the original vocabulary and model scale is essential. By maintaining the full 300M-parameter architecture and vocabulary through a Continued Pre-Training (CPT) strategy~\cite{gururangan-etal-2020-dont}, we mitigate the risk of catastrophic forgetting regarding the model's foundational multilingual capabilities while capturing the dense terminological and structural complexities inherent to technical corpora~\cite{legal-bert}.

Despite these developments, a significant challenge persists in reconciling broad multilingual coverage with the rigorous requirements of high-stakes specialization. While specialized encoders have been developed for the biomedical~\cite{ClinicalModernBERT,bioclinicalModernBERT} and legal~\cite{legal-bert} domains, these models often remain decoupled from the architectural improvements seen in modern general-purpose encoders. We argue that the optimal path to specialization is context-dependent. For regional languages such as Spanish and Catalan, efficiency is best achieved through vocabulary adaptation~\cite{flor} and language-specific data mining~\cite{RobertaCA, roberta-base-bne, rigoberta}, allowing for more compact, task-optimized footprints. Conversely, for knowledge-dense and terminologically complex domains like law and biomedicine, preserving the original model scale and broad vocabulary is essential. By employing a Continued Pre-Training (CPT) strategy~\cite{gururangan-etal-2020-dont} on a 300M-parameter architecture, we preserve foundational multilingual capabilities while enabling the model to internalize the dense technical notation and structural complexities characteristic of legal and biomedical corpora.

In this work, we introduce MrBERT, a family of 150M and 300M-parameter encoders built on the ModernBERT architecture and pre-trained on 35 languages and code. Through targeted adaptation strategies, we derive computationally efficient English–Spanish and Catalan–English variants via vocabulary specialization, and develop domain-adapted models for biomedical and legal contexts through continued pre-training (CPT). Our models achieve state-of-the-art performance on the CLUB (Catalan) and EvalES (Spanish) benchmarks~\cite{CLUB, evalES}, while demonstrating robust performance across domain-specific text classification, named-entity recognition, and retrieval tasks.

Furthermore, we address the challenge of representation efficiency through MRL~\cite{matformer, matryoshkaRL}. In production environments, particularly in specialized fields like law or biomedicine, retrieval systems must frequently balance high-resolution accuracy with the latency constraints of massive databases. We provide a rigorous empirical study of two architectural approaches to MRL: MLP-based projection and multi-head attention groupings. This analysis explores the trade-offs between computational latency and resolution, providing a blueprint for deploying encoders across varied hardware constraints.

Our contributions are as follows: 
\begin{itemize} 
    \item \textbf{MrBERT Foundation}: A 300M-parameter multilingual model built on the ModernBERT architecture that demonstrates competitive performance across multilingual benchmarks while serving as a robust base for targeted adaptation. 
    \item \textbf{Language Adaptation}: Leveraging vocabulary adaptation to provide state-of-the-art, computationally efficient alternatives for Spanish and Catalan NLU. 
    \item \textbf{Domain Specialization}: A suite of models adapted for Legal and Biomedical domains via CPT. These models outperform existing specialized encoders.
    \item \textbf{Matryoshka Analysis}: A systematic investigation into architectural variants for flexible embeddings, offering insights into optimizing modern encoders for varied retrieval and deployment constraints.
\end{itemize}

%Through a controlled comparison framework based on standardized continual pre-training, we show that MrBERT provides a practical and scalable approach to training multilingual and domain-adapted encoder models, offering an alternative to both general-purpose multilingual encoders and narrowly specialized models.

By synthesizing modern architecture with targeted adaptation strategies, MrBERT provides a scalable framework that bridges the gap between general-purpose multilingualism and narrow domain expertise.

\section{Related Work}
Masked, bidirectional encoders remain the core paradigm for dense representation learning. Since BERT \cite{BERT}, the field has progressed through RoBERTa \cite{liu2019roberta} to large-scale multilingual models like XLM-RoBERTa and DeBERTa \cite{xlm-roberta, deberta}. Recently, encoder-only architectures have been revitalized through modern recipes like ModernBERT \cite{modernbert} and Ettin \cite{ettin}, incorporating RoPE, GeGLU activations, and unpadding strategies for long contexts and memory efficiency.

\paragraph{The `Extraction vs. Pre-training' Debate}
A central debate concerns whether to extract encoders from hybrid architectures or train from scratch. EmbeddingGemma \cite{gemma} shows that adapting pre-trained weights from encoder-decoder models can be compute-efficient, while Ettin \cite{ettin} demonstrates that encoders pre-trained with dedicated bidirectional objectives consistently outperform extracted counterparts on NLU and retrieval tasks. This motivates our choice to build MrBERT as a natively trained ModernBERT-based encoder.

\paragraph{Adaptation and Multilinguality}
Effective retrieval requires balancing cross-lingual transfer with local efficiency. While mGTE \cite{mgte} emphasizes long-context retrieval objectives, language-specific models like FLOR \cite{flor} and Spanish/Catalan variants \cite{RobertaCA, roberta-base-bne} show that targeted vocabulary adaptation outperforms generic multilingual tokenizers. Similarly, continued pre-training on domain-specific corpora, as seen in BioClinicalModernBERT \cite{bioclinicalModernBERT} and Legal-BERT \cite{legal-bert}, remains the standard for capturing dense terminological complexities. We bridge these trends by applying language adaptation to a modern, long-context architecture.

\paragraph{Flexible Representation Learning}
Matryoshka Representation Learning (MRL) \citep{matryoshkaRL} enables ``nested" embeddings with semantic consistency across dimensions. While early work focused on MLP projections, FlexTron \citep{flextron} shifted focus to attention mechanisms, motivated by observations that attention heads are often redundant \citep{multihead_pruning} or specialized \citep{nam2025causalheadgatingframework, iti, memat}. Applying matryoshka principles to attention heads, a bottleneck scaling quadratically with sequence length, enables efficient adaptive computation as demonstrated by HydraViT \citep{hydraVIT} and ThinkingViT \citep{thinkingVIT}. We systematically compare MLP-based and attention-based matryoshka variants, showing that while MLP configurations retain a slight performance edge, attention-based variants provide superior inference-time efficiency for production deployment.

\section{Pre-training and Adaptation}
\subsection{Data}

The training process was conducted in three separate stages: large-scale Pre-Training, followed by Language Adaptation, and concluding with Domain Adaptation. Across all phases, we applied a standardized curation pipeline to ensure data quality, as described below.

\begin{figure*}
    \centering
    \includegraphics[width=0.9\linewidth]{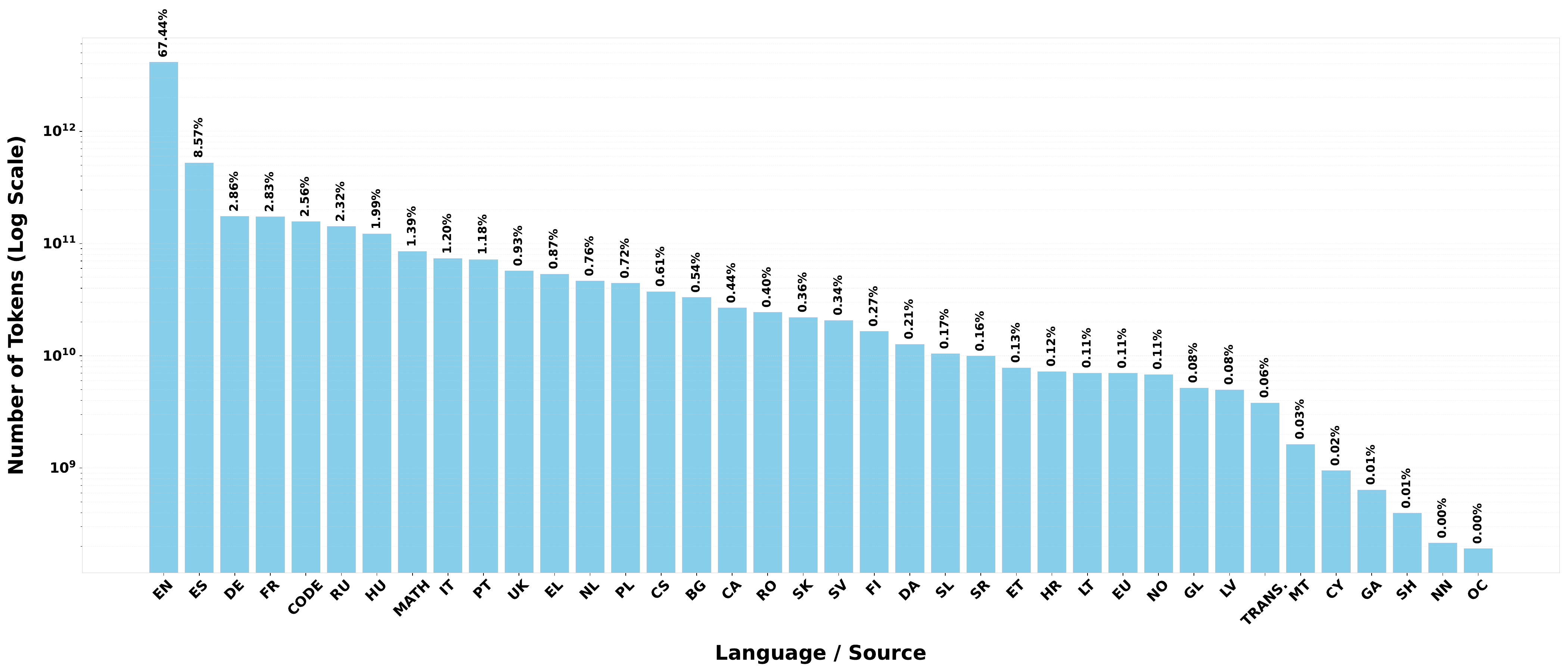}
    \caption{Token distribution per language for the Pre-Training phase. The table is shown in logarithmic format for visualization purposes.}
    \label{fig:pretrain-dist}
\end{figure*}

\paragraph{Pre-Training} Figure~\ref{fig:pretrain-dist} shows the number of tokens per language used during the pre-training phase. A comprehensive list of data sources used throughout training is provided in Appendix~\ref{app:data_sources} and the exact values for each language are presented in Appendix~\ref{app:lang_sources}.

All datasets were processed using the CURATE~\cite{palomar2024curated} pipeline. We applied document-level exact deduplication across the corpus and removed documents with quality score below 0.2 provided by the CURATE pipeline. For datasets providing intrinsic quality scores (e.g., FineWeb-Edu, FineWeb2-HQ), documents were sorted in descending score order and only the highest-scoring portions were retained.

For parallel translated data, following prior work, we concatenate source and target pairs and insert the special token \texttt{<|translation|>} between them \cite{parallelPretraining,eurobert}. This format allows the model to jointly learn translation and multilingual alignment during pre-training.

\paragraph{Language Adaptation}

After multilingual pre-training using the Salamandra tokenizer \cite{salamandra}, we adapt the vocabulary of the tokenizer and perform a second training stage focused on language adaptation. In this phase, we restrict training to bilingual mixtures with equal sampling weights (50\%--50\%) between English and a target language. As in pre-training, all datasets undergo document-level exact deduplication and Curate-based filtering, and full dataset details are deferred to Appendix~\ref{app:data_sources}.

We train two language-adapted variants:

\begin{itemize}
    \item \textbf{EN--ES adaptation:} A total of 615B tokens, sampled with a 50\% English and 50\% Spanish mixture.
    \item \textbf{EN--CA adaptation:} A total of 47.4B tokens, sampled with a 50\% English and 50\% Catalan mixture.
\end{itemize}

For Spanish adaptation, English data is primarily drawn from high-quality subsets of large-scale English corpora, complemented with general-domain data to preserve linguistic diversity. For Catalan adaptation, a larger proportion of Catalan-specific corpora is used to compensate for the smaller availability of high-quality Catalan data, while maintaining a balanced bilingual mixture. 

\paragraph{Domain Adaptation}
To maintain broad language coverage and prevent the model from specializing too early on a restricted vocabulary, we perform domain adaptation directly on the multilingual base model rather than the language-adapted versions. This ensures that the model learns domain-specific knowledge before the representation space is narrowed to a specific language pair.

Domain adaptation is carried out through continual pre-training on domain-specific corpora. While the process supports multiple languages, we intentionally focus the data mixture on English and Spanish to match our evaluation priorities and data availability.

We focus on two target domains:

\begin{itemize}
\item \textbf{Legal Adaptation:} The model is trained on 9B tokens, consisting of 79.5\% English and 20.5\% Spanish data. Because the dataset is relatively small and the validation loss continued to improve, we trained for 10 epochs. We found no evidence of overfitting during this stage.

\item \textbf{Biomedical Adaptation:} This stage uses a larger 24B-token corpus, primarily composed of English (84.7\%) and Spanish (14.8\%) data. We also include small amounts of German (0.18\%), Italian (0.11\%), and French (0.11\%) to maintain a degree of multilingual breadth. Based on validation loss trends and the scale of the data, we trained for 2 epochs to ensure stable generalization.
\end{itemize}

This approach is designed to enable domain specialization, allowing us to systematically study domain effects specifically in English and Spanish.

\paragraph{Classification of targeted domain instances} While the datasets used for domain adaptation are broadly categorized into Legal and Biomedical domains, a manual qualitative analysis revealed significant internal variance. Many datasets contain ``noisy" instances from unrelated subdomains or exhibit inherent thematic overlap. To ensure high-quality domain alignment, we employed NVIDIA's multilingual domain classifier\footnote{\href{https://huggingface.co/nvidia/multilingual-domain-classifier}{https://huggingface.co/nvidia/multilingual-domain-classifier}} to filter and refine the instances.

\paragraph{Domain Mapping and Selection Logic}
The classifier categorizes text into 26 distinct classes (see Appendix~\ref{app:classifier_categories} for the full list). To align these with our research objectives, we established the following mapping:

\begin{itemize}
    % \item \textbf{Scientific} Domain: Mapped from the ``Science'' class.
    
    \item \textbf{Biomedical Domain:} Mapped from the ``Health'' class.

    \item \textbf{Legal Domain:} Mapped from the ``Law and Government'' class.
\end{itemize}

We employed a Top-1 selection strategy, where an instance was assigned to a target domain only if the corresponding mapped class received the highest probability.

\subsection{Pre-Training Settings}
Following the ModernBERT training recipe \cite{modernbert}, MrBERT is pre-trained using a three-stage strategy with a Warmup–Stable–Decay (WSD) learning-rate schedule, optimized via StableAdamW.

\begin{itemize}
\item \textbf{Short-context pre-training}: Sequence length of 1,024, trained on 5.5T tokens.
\item \textbf{Long-context adaptation}: The RoPE scaling parameter in global attention layers is increased to 160{,}000, with training continued on 500B tokens.
\item \textbf{Annealing}: The sequence length is fixed at 8,192, and a $1-sqrt$ learning-rate decay \cite{scaling_laws_lr} is applied over 100B tokens, progressively emphasizing higher-quality data to improve final model performance.
\end{itemize}

We adopt the original ModernBERT framework\footnote{\href{https://github.com/AnswerDotAI/ModernBERT}{https://github.com/AnswerDotAI/ModernBERT}}. During preprocessing, we insert explicit document separators by concatenating end-of-sequence (EOS) and beginning-of-sequence (BOS) tokens between documents and we further adjust the padding and attention masking logic to prevent any attention across document boundaries under this packing scheme. This preprocessing strategy is applied consistently across all pre-training stages and subsequent adaptations. 

\subsection{Language and Domain Specialization}

\paragraph{Vocabulary Adaptation and Initialization.} We trained dedicated Spanish and Catalan tokenizers with a vocabulary size of $V \approx 50,000$ and adapted our multilingual encoder following the strategy proposed by \cite{vocab_adptation_ogs}. This strategy adapts the embedding layer to the new tokenizer by reusing embeddings for shared tokens. Analysis of the vocabulary intersection reveals that adapting the multilingual base to Spanish retains a $64.24\%$ token overlap. Although the overlap between Spanish and Catalan is significantly lower ($32.15\%$), empirical results indicate that initializing the Catalan adaptation with Spanish-adapted weights yields superior validation perplexity. This suggests that the model effectively leverages shared Romance morphological features, accelerating the alignment of the extended embedding space. We tailored the optimization for each language: for Spanish, we employed a Warmup-Stable-Decay (WSD) schedule, while for Catalan, we utilized a Warmup + Cosine Decay approach.

\paragraph{Domain Adaptation.} For the subsequent specialization into legal and biomedical domains, we also adopted a Warmup + Cosine Decay scheduler. To optimize this transition, we conducted a hyperparameter sweep over peak learning rates and epoch counts, selecting the checkpoints that achieved the global minimum in validation cross-entropy for final evaluation.

A detailed list of hyperparameters for all models is provided in Appendix~\ref{app:hyper}.

\section{Efficient Representations: Matryoshka Architectures}

Given the widespread adoption of MRL \cite{matryoshkaRL,matformer} in retrieval models \cite{mgte,gemma}, we extend its application to encoder-based architectures. Following the methodology of \cite{flextron}, we study matryoshka along two architectural dimensions: attention heads and the intermediate MLP projections. To evaluate these two variants independently, we replace the standard annealing phase in the multilingual model with a combined annealing+matryoshka phase.

This integrated phase acts as a curriculum learning strategy, closely related to the Sequential Matryoshka Representation Learning (SMRL) framework, and helps reduce gradient variance by stabilizing shared parameters across multiple granularities during the final stages of pre-training \cite{curriculum_mat}. Such strategies have become increasingly common in high-performance embedding pipelines (e.g., mGTE), where maintaining semantic consistency of hierarchical representations across languages is critical.

\newpage
\section{Evaluation}
\subsection{Overview}

\paragraph{Multilingual Evaluation}
We evaluate multilingual performance using XTREME \cite{xtreme}. To ensure a fair comparison, we modify the original evaluation protocol, as the native framework uses model-specific, hard-coded learning rates. Instead, we fine-tune each model exclusively on English while sweeping over five learning rates. The optimal learning rate is selected based on validation performance, and final results are reported on the test split. We omit evaluations on Tatoeba and BUCC \cite{tatoeba, bucc}, as retrieval is extensively covered in our domain-specific experiments using MTEB \cite{mteb}. Since our model does not cover all XTREME languages, we restrict evaluation to the languages included in our training data.

\paragraph{Monolingual Evaluation}
For monolingual evaluation, we use CLUB (Catalan) \cite{CLUB} and EvalES (Spanish) \cite{evalES}. Following the XTREME setup, we perform a learning rate sweep over three values and report test results corresponding to the model achieving the best validation performance.

\paragraph{Domain-Specific Evaluation}
We evaluate domain-specific performance using a subset of MTEB \cite{mteb} tasks covering the legal and biomedical domains. Following \cite{modernbert}, we adopt a ColBERT-style training approach \cite{colbert}, distilling knowledge from a teacher model by minimizing the KL divergence between normalized teacher and student similarity scores. Models are trained on 810k samples from MS MARCO\footnote{\href{https://huggingface.co/datasets/lightonai/ms-marco-en-bge}{https://huggingface.co/datasets/lightonai/ms-marco-en-bge}} \cite{msmarco}, using teacher scores generated by BGE-M3 \cite{m3Embedding}. Training is conducted with the PyLate library \cite{pylate}. We reserve 1\% of the training data as a validation set for selecting the best model across four learning rates. The model with the optimal validation performance is then evaluated on specific tasks of MTEB.

% \footnote{
% CANTEMIST~\cite{cantemist}, PHARMACONER~\cite{pharmaconer}, and DISTEMIST~\cite{distemist} (using the Hugging Face release
% \texttt{BSC-NLP4BIA/bsc-bio-distemist-ner} for ease of access and reproducibility).
% }
% \footnote{EURLEX \cite{eurlex}.}
To further enhance the coverage of Spanish evaluations in domain-specific scenarios, we incorporate three biomedical named-entity recognition datasets \cite{cantemist, pharmaconer, distemist}, one legal text classification dataset \cite{eurlex}, and create two novel Spanish datasets\footnote{Both novel datasets are constructed from openly licensed data.}:

\begin{itemize}
    \item \textbf{Legal:} \textbf{LexBOE\footnote{\url{https://huggingface.co/datasets/BSC-LT/LexBOE}}} is a Spanish legal text classification dataset built from articles published in the \emph{Boletín Oficial del Estado} between 2022 and 2024, extracted via the official BOE API\footnote{\url{https://www.boe.es/datosabiertos/api/api.php}}. Documents are assigned to one of 14 legal labels obtained through manual unification of the original metadata. The texts are pseudo-anonymized using semantically and formally equivalent replacements to preserve linguistic structure.
    \item \textbf{Biomedical:} \textbf{AbSanitas\footnote{\url{https://huggingface.co/datasets/BSC-LT/AbSanitas}}} is a Spanish biomedical information retrieval dataset built from biomedical abstracts collected from the RECOLECTA dataset (see Appendix~\ref{app:recolecta} for further details). Each document is associated with two distinct synthetically generated queries, validated through LLM-as-a-Judge.\footnote{Queries were generated using DeepSeek V3 \cite{deepseekv3} and validated using Qwen3-32B as an LLM-as-a-Judge \cite{qwen3}.}
\end{itemize}

Since our domain evaluation setup is bilingual while most domain adaptations are English-centric, we report English-only scores separately to enable a fair comparison between our model and English-only variants. Spanish-only models such as Rigoberta \cite{rigoberta} are excluded, as the retrieval adaptations rely on an English-only dataset, which led to unstable training dynamics and degraded task performance for these models. Finally, given that our study focuses on assessing the ability of base domain models under identical fine-tuning conditions, we exclude models such as EmbeddingGemma and mGTE, whose architectures are natively designed for retrieval and are fundamentally different from ColBERT-style approaches.

\subsection{Results}

\begin{table*}[t!]
\centering
\begin{tabular}{cccccc}
\\ \hline
\textbf{task} & \textbf{\begin{tabular}[c]{@{}c@{}}xlm-roberta-base\\ (279M)\end{tabular}} & \textbf{\begin{tabular}[c]{@{}c@{}}mRoBERTa\\ (283M)\end{tabular}} & \textbf{\begin{tabular}[c]{@{}c@{}}mmBERT\\ (308M)\end{tabular}} & \textbf{\begin{tabular}[c]{@{}c@{}}mGTE\\ (306M)\end{tabular}} & \textbf{\begin{tabular}[c]{@{}c@{}}MrBERT\\ (308M)\end{tabular}} \\ \hline
UD-POS (F1)         & \textbf{85.55}                                                             & {\ul 85.36}                                                        & 84.33                                                            & 82.50                                                          & 83.74                                                            \\
PANX (F1)          & 73.69                                                                      & \textbf{75.65}                                                     & {\ul 73.89}                                                      & 73.05                                                          & 72.06                                                            \\
XNLI (Acc.)          & 78.25                                                                      & 79.09                                                              & {\ul 80.54}                                                      & 77.90                                                          & \textbf{81.26}                                                   \\
PAWS-X (Acc.)        & 89.50                                                                      & 90.36                                                              & \textbf{92.34}                                                   & 89.55                                                          & {\ul 91.32}                                                      \\
TyDiQA (F1)        & {\ul 56.41}                                                                & 53.96                                                              & \textbf{63.95}                                                   & 51.07                                                          & 56.34                                                            \\
MLQA (F1)          & 68.91                                                                      & 68.67                                                              & \textbf{71.48}                                                   & 68.05                                                          & {\ul 70.67}                                                      \\
XQuAD (F1)        & 75.61                                                                      & 75.45                                                              & {\ul 77.79}                                                      & 74.37                                                          & \textbf{77.91}                                                   \\ \hline
Average       & 75.42                                                                      & 75.51                                                              & \textbf{77.76}                                                   & 73.78                                                          & {\ul 76.19}                                                      \\ \hline                           
\end{tabular}\caption{Multilingual performance on XTREME benchmark tasks. Models fine-tuned on English data with learning rates selected by validation performance. }\label{tab:xtreme-results}
\end{table*}

\begin{table*}[t!]
\centering
\begin{tabular}{cccccccc}
\hline
\textbf{tasks} & \textbf{\begin{tabular}[c]{@{}c@{}}xlm-roberta\\ -base (279M)\end{tabular}} & \textbf{\begin{tabular}[c]{@{}c@{}}mRoBERTa\\ (283M)\end{tabular}} & \textbf{\begin{tabular}[c]{@{}c@{}}mmBERT\\ (308M)\end{tabular}} & \textbf{\begin{tabular}[c]{@{}c@{}}mGTE\\ (306M)\end{tabular}} & \textbf{\begin{tabular}[c]{@{}c@{}}MrBERT\\ (308M)\end{tabular}} & \textbf{\begin{tabular}[c]{@{}c@{}}MrBERT-es\\ (150M)\end{tabular}} \\ \hline
UD-POS-es (F1)                                                              & 99.01                                                                       & 99.03                                                              & \textbf{99.09}                                               & 98.92                                                          & 99.06                                                            & {\ul 99.08}                                                         \\
CoNLL-NERC-es (F1)                              & 86.91                                                                       & \textbf{87.77}                                                        & 87.01                                                            & 86.96                                                          & 87.42                                                            & {\ul 87.77 }                                                             \\
STS-es (Pearson)  & 80.88                                                                       & 79.69                                                              & 82.88                                                            & {\ul 84.52}                                                    & 84.18                                                            & \textbf{85.23}                                                  \\
PAWS-X-es (Acc.)  &  90.35                                                                       & 91.30                                                              & {\ul 91.35}                                                      & 89.70                                                          & 91.25                                                          & \textbf{91.90}                                                  \\
MlDoc (Acc.) & 47.67                                                                       & 91.28                                                              & 95.10                                                            & \textbf{96.13}                                             & 95.28                                                      & {\ul 95.55}                                                               \\
Massive (Acc.)      & 21.89                                                                       & 86.45                                                              & 86.79                                                            & {\ul 87.19}                                             & \textbf{87.46}                                                            & 87.05                                                         \\
SQAC (F1) & 74.48                                                                       & 77.03                                                              & 79.79                                                            & 76.78                                                          & {\ul 81.96}                                                      & \textbf{82.19}                                                  \\ \hline
Average    & 71.60                                                                      & 87.51                                                            & 88.86                                                            & 88.60                                                         & {\ul 89.52}                                                      & \textbf{89.83}                                                  \\ \hline
\end{tabular}
\caption{Performance on Spanish language tasks from the EvalES benchmark. } \label{tab:es_eval-results}
\end{table*}

\begin{table*}[t!]
\centering
\begin{tabular}{ccccccccc}
\hline
\textbf{tasks} & \textbf{\begin{tabular}[c]{@{}c@{}}xlm-roberta\\ -base (279M)\end{tabular}} & \textbf{\begin{tabular}[c]{@{}c@{}}mRoBERTa \\ (283M)\end{tabular}} & \textbf{\begin{tabular}[c]{@{}c@{}}roberta-ca \\ (125M)\end{tabular}} & \textbf{\begin{tabular}[c]{@{}c@{}}mmBERT \\ (308M)\end{tabular}} & \textbf{\begin{tabular}[c]{@{}c@{}}mGTE \\ (306M)\end{tabular}} & \textbf{\begin{tabular}[c]{@{}c@{}}MrBERT \\ (308M)\end{tabular}} & \textbf{\begin{tabular}[c]{@{}c@{}}MrBERT-ca \\ (150M)\end{tabular}} \\ \hline
AnCora-ca-ner (F1)       & 87.61                                                                       & {\ul 88.33}                                                         & \textbf{89.70}                                                    & 88.14                                                             & 87.20                                                           & 87.32                                                             & 88.04                                                                \\
AnCora-ca-pos (F1)       & 98.91                                                                       & 98.98                                                               & 99.00                                                                 & {\ul 99.01}                                                             & 98.77                                                         & 99.01                                                             &  \textbf{99.03}                                                          \\ 
STS-ca (Pearson)   & 74.67                                                                       & 79.52                                                               & 82.99                                                                 & {\ul 83.16}                                                       & 78.65                                                           & 83.00                                                             & \textbf{85.42}                                                       \\
TeCla (Acc.)      & 72.57                                                                       & 72.41                                                               & 72.81                                                                 & 74.11                                                             & {\ul 74.68}                                                     & 73.79                                                             & \textbf{74.97}                                                   \\
TECA (Acc.)      & 79.59                                                                       & 82.38                                                               & 82.14                                                                 & 83.18                                                             & 79.40                                                           & {\ul 84.03}                                                       & \textbf{86.92}                                                   \\
ViquiQuAD (F1) & 86.93                                                                       & 87.86                                                               & 87.31                                                                 & \textbf{89.86}                                                & 86.78                                                           & 89.25                                                             & {\ul 89.59}                                                          \\
XQuAD (F1)     & 69.69                                                                       & 69.40                                                               & 70.53                                                                 & 73.88                                                             & 69.27                                                           & {\ul 73.96}                                                       & \textbf{74.47}                                                   \\ \hline
Average        & 81.42                            & 82.70                    & 83.50                      & {\ul 84.48}            & 82.09                & 84.34                  & \textbf{85.49}               
\\ \hline
\end{tabular}
\caption{Performance on Catalan language tasks from the CLUB benchmark.
}\label{tab:club-results}
\end{table*}

\paragraph{Multilingual and Monolingual Results} The evaluation across XTREME (Table~\ref{tab:xtreme-results}), Spanish (Table~\ref{tab:es_eval-results}), and Catalan (Table~\ref{tab:club-results}) benchmarks reveals a consistent performance hierarchy favoring the MrBERT and mmBERT models. In the broad multilingual setting, mmBERT establishes a strong baseline with an average score of 77.76, notably outperforming xlm-roberta-base (75.42) in dense linguistic tasks like Question Answering. However, the most significant gains are observed in the language-specific models.

\begin{table*}[t!]
\centering
\begin{tabular}{cccccccc}
\hline
\textbf{Task Name} & \textbf{Task Type}                                                               & \textbf{\begin{tabular}[c]{@{}c@{}}mmBERT\\ (308M)\end{tabular}} & \textbf{\begin{tabular}[c]{@{}c@{}}MrBERT\\ (308M)\end{tabular}} & \textbf{\begin{tabular}[c]{@{}c@{}}MrBERT\\ -es (150M) \end{tabular}} & \textbf{\begin{tabular}[c]{@{}c@{}}BioClinical-\\ MdnBERT \\ (150M)\end{tabular}} & \textbf{\begin{tabular}[c]{@{}c@{}}Clinical\\ MdnBERT \\ (137M)\end{tabular}} & \textbf{\begin{tabular}[c]{@{}c@{}}MrBERT\\ -biomed \\ (308M) \end{tabular}} \\ \hline
\begin{tabular}[c]{@{}c@{}}bsc-bio\\ -distemist-ner (ES)\end{tabular} & NER       & {\ul 78.00}    & 77.84          & \textbf{78.07}       & 75.45                  & 70.22                & 77.93                \\
cantemist (ES)                                                        & NER       & \textbf{78.03} & 68.73          & {\ul 73.40}          & 66.68                  & 30.91                & 70.78                \\
pharmaconer (ES)                                                      & NER       & {\ul 89.66}    & 88.58          & 88.97                & 87.66                  & 81.69                & \textbf{89.92}       \\
AbSanitas (ES)                                                        & Retrieval & 34.68          & 34.16          & \textbf{53.49} & 30.41                  & 18.08                & {\ul 51.01} \\ \hline
R2Med (EN)                                                            & Retrieval & \textbf{10.87} & {\ul 10.15}    & 8.65           & 9.97                   & 5.91                 & 9.76                 \\
SciDocs (EN)                                                          & Retrieval & {\ul 10.00}    & 9.75           & 9.90                 & 9.33                   & 3.64                 & \textbf{10.05}       \\
SciFact (EN)                                                          & Retrieval & \textbf{32.35} & 31.08          & 31.46                & {\ul 32.07}            & 20.34                & 30.25                \\
TREC-COVID (EN)                                                      & Retrieval & 30.77          & \textbf{49.53} & 37.51                & 46.08                  & 23.88                & {\ul 48.76}          \\ \hline
Average (EN)                                                          & All Tasks & 21.00          & \textbf{25.13} & 21.88                & 24.36                  & 13.44                & {\ul 24.71}          \\ \hline
Average (EN + ES)                                                     & All Tasks & 45.55          & 46.23          & {\ul 47.68}          & 44.71                  & 31.83                & \textbf{48.56}       \\ \hline
\end{tabular}
\caption{Biomedical domain evaluation on Spanish and English retrieval and classification tasks. The R2Med score is reported as the average over the bioinformatics, biology, and clinical subsets. Retrieval performance is measured using nDCG@10, while NER is evaluated using F1.} \label{tab:mteb-biomed-results}
\end{table*}

\begin{table*}[t!]
\centering
\begin{tabular}{ccccccc}
\hline
\textbf{Task Name} &  \textbf{Task Type}      & \textbf{\begin{tabular}[c]{@{}c@{}}mmBERT\\ (308M)\end{tabular}} & \textbf{\begin{tabular}[c]{@{}c@{}}MrBERT\\ (308M)\end{tabular}} & \textbf{\begin{tabular}[c]{@{}c@{}}MrBERT-es\\ (150M)\end{tabular}} & \textbf{\begin{tabular}[c]{@{}c@{}}legal-bert-\\ base-uncased\\ (110M)\end{tabular}} & \textbf{\begin{tabular}[c]{@{}c@{}}MrBERT\\ -legal\\ (308M)\end{tabular}} \\ \hline
LexBOE (ES)                                                                & Text Classification & 96.84          & {\ul 97.02} & \textbf{97.28} & 95.36                   & 96.80          \\
\begin{tabular}[c]{@{}c@{}}small-spanish\\ -legal-dataset (ES)\end{tabular} & Retrieval           & {\ul 42.58}    & 40.78       & \textbf{46.92} & 19.79                   & 38.75          \\ \hline
EURLEX (EN)                                                                & Text Classification & \textbf{97.43} & 97.40       & 97.41          & {\ul 97.42}             & 97.33          \\
AILAStatutes (EN)                                                          & Retrieval           & {\ul 14.31}    & 13.90       & 12.28          & 13.49                   & \textbf{16.33} \\
legal\_summarization (EN)                                                  & Retrieval           & 53.33          & {\ul 53.84} & 46.41          & 52.40                   & \textbf{55.05} \\
LegalBench (EN)                                                            & Retrieval           & {\ul 60.15}    & 58.88       & 58.26          & \textbf{63.42}          & 58.04          \\
NanoTouche2020 (EN)                                                      & Retrieval           & 34.03          & {\ul 44.15} & 31.18          & 34.48                   & \textbf{44.74} \\ \hline
Average (EN)                                                           & All Tasks           & 51.85          & {\ul 53.63} & 49.11          & 52.24                   & \textbf{54.30} \\ \hline
Average (EN + ES)                                                      & All Tasks           & 56.95          & {\ul 58.00} & 55.68          & 53.77                   & \textbf{58.15} \\ \hline
\end{tabular}\caption{Legal domain evaluation on Spanish and English retrieval and classification tasks. The LegalBench score is reported as the average over the consumer contracts and corporate lobbying subsets. Retrieval tasks are evaluated using nDCG@10, while text classification tasks are evaluated using accuracy. }\label{tab:mteb-legal-results}
\end{table*}

While multilingual MrBERT performs robustly, our specialized MrBERT-es and MrBERT-ca models achieve State-of-the-Art (SOTA) results in Spanish (89.83) and Catalan (85.49), respectively. Remarkably, these 150M-parameter models outperform their larger 308M-parameter parent versions despite having half the parameter count. This performance gap is particularly evident in Spanish classification tasks ($MlDoc$ and $Massive$), where standard multilingual models like xlm-roberta-base exhibit significant instability.

By reducing the parameter count while maintaining or exceeding the accuracy of larger models, these versions provide a superior balance of computational efficiency and performance, making them highly suitable for resource-constrained production environments.

\subsection{Domain-Specific Results}

\paragraph{Biomedical Domain}
Table \ref{tab:mteb-biomed-results} presents evaluation results across biomedical tasks. MrBERT-biomed achieves the best overall performance, substantially outperforming existing domain-specific baselines. The improvement is most pronounced on the Spanish retrieval task AbSanitas, where domain adaptation yields significant gains over general multilingual models. We observe heterogeneous performance patterns in Spanish NER tasks: mmBERT remain highly competitive on cantemist, while MrBERT-biomed demonstrates its advantage on pharmaconer. On English biomedical tasks, MrBERT achieves the strongest average performance, while existing specialized models like Clinical ModernBERT show substantially weaker results, highlighting the effectiveness of our training approach.

\paragraph{Legal Domain}
Table \ref{tab:mteb-legal-results} shows evaluation results for legal tasks. MrBERT-legal (308M) achieves the best overall performance with an average of 58.15, with consistent improvements on retrieval tasks. MrBERT-es (150M) demonstrates exceptional performance on Spanish legal tasks despite having half the parameters, achieving 97.28 on LexBOE classification and 46.92 on small-spanish-legal-dataset retrieval. Text classification tasks show high performance across all models, while retrieval tasks reveal more substantial differences where domain adaptation provides clear benefits.

\subsection{Matryoshka Results}

\begin{figure}[h]
    \centering
    \includegraphics[width=1\linewidth]{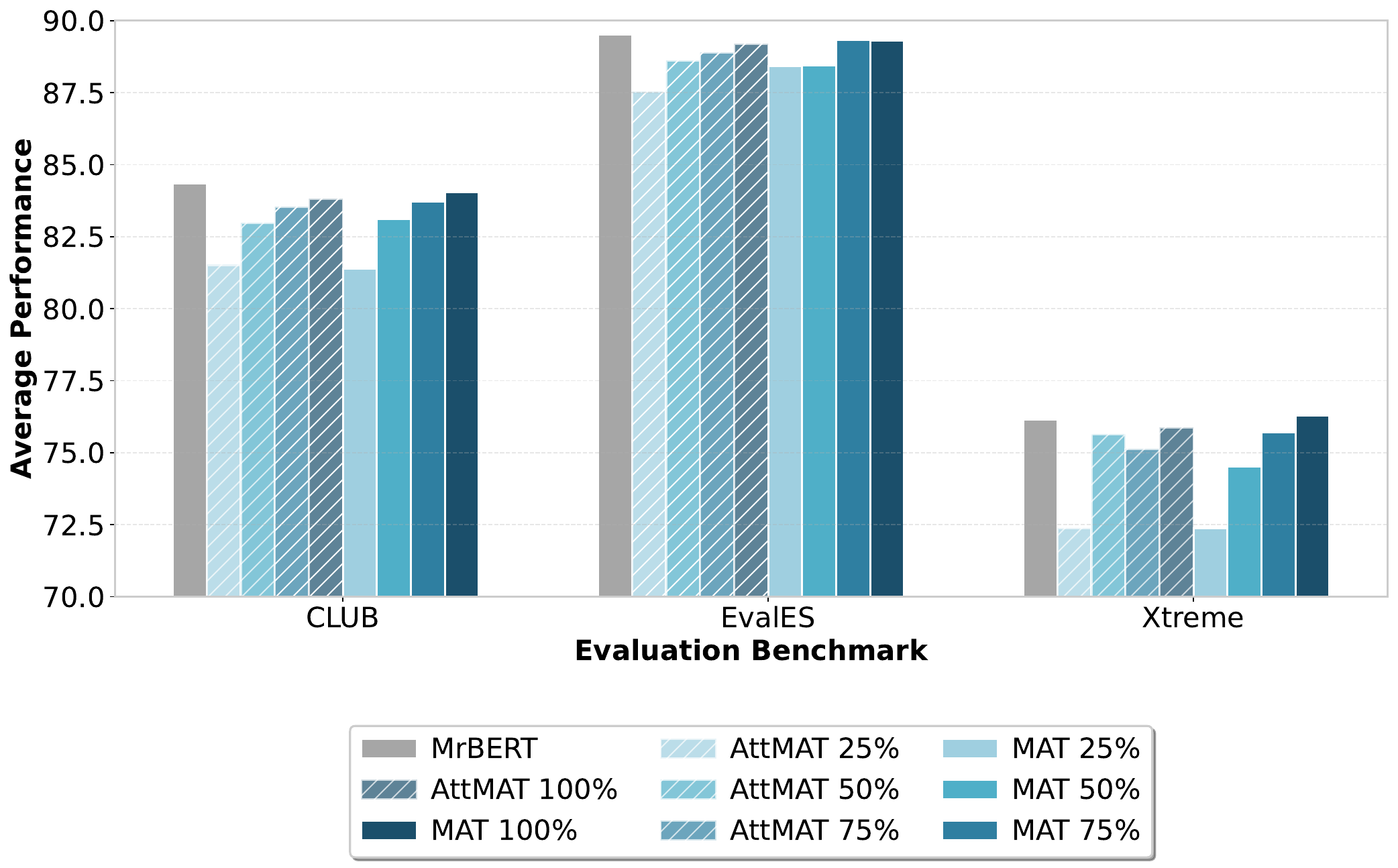}
    \caption{MrBERT performance across XTREME, CLUB, and EvalES benchmarks comparing AttMAT (attention head pruning), MAT (MLP hidden size reduction), and standard models (100\%). Only average scores shown.}
    \label{fig:benchmarks_math}
\end{figure}

\begin{figure}[h]
    \centering
    \includegraphics[width=1\linewidth]{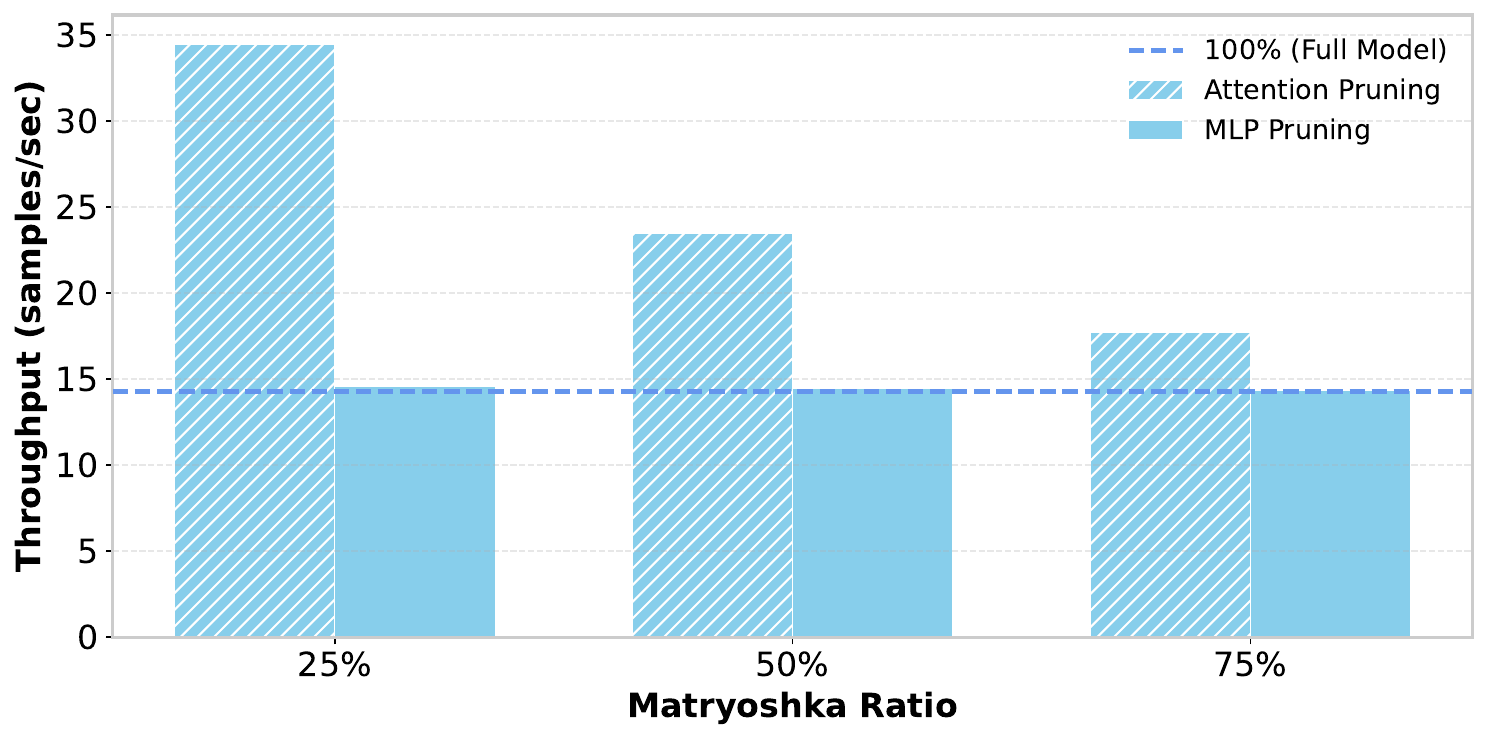}
    \caption{Inference throughput of matryoshka variants (sequence length: 8,192 tokens).}
    \label{fig:matr}
\end{figure}

As shown in Figure \ref{fig:benchmarks_math}, the MLP-based matryoshka variant yields slightly better downstream performance. This observation is consistent with prior work \cite{matformer}, which attributes the robustness of sliced MLP representations to their high parameter density and expressive capacity. However, when considering inference-time memory footprint and latency, Figure \ref{fig:matr} shows that attention-head matryoshka offers superior efficiency. Since our primary objective is to obtain the fastest possible models, we therefore adopt the attention-based matryoshka configuration in our final models. This choice is further supported by recent scalable and “thinking-based” architectures such as ThinkingViT and HydraViT \cite{hydraVIT,thinkingVIT}, which emphasize attention-head elasticity as a key mechanism for hardware-aware efficiency. Detailed results for all matryoshka experiments are provided in Appendix \ref{app:matry}.

To further evaluate this approach, we adopt the same data and training configuration used in the language and domain adaptation experiments and adapt the models using the matryoshka scheme. Due to the substantial computational budget allocated to the Spanish dataset, we replicate this setup by enabling matryoshka only during the annealing phase. For all other adaptation settings, matryoshka is applied throughout the entire adaptation process.  

\begin{figure}[h]
    \centering
    \includegraphics[width=1\linewidth]{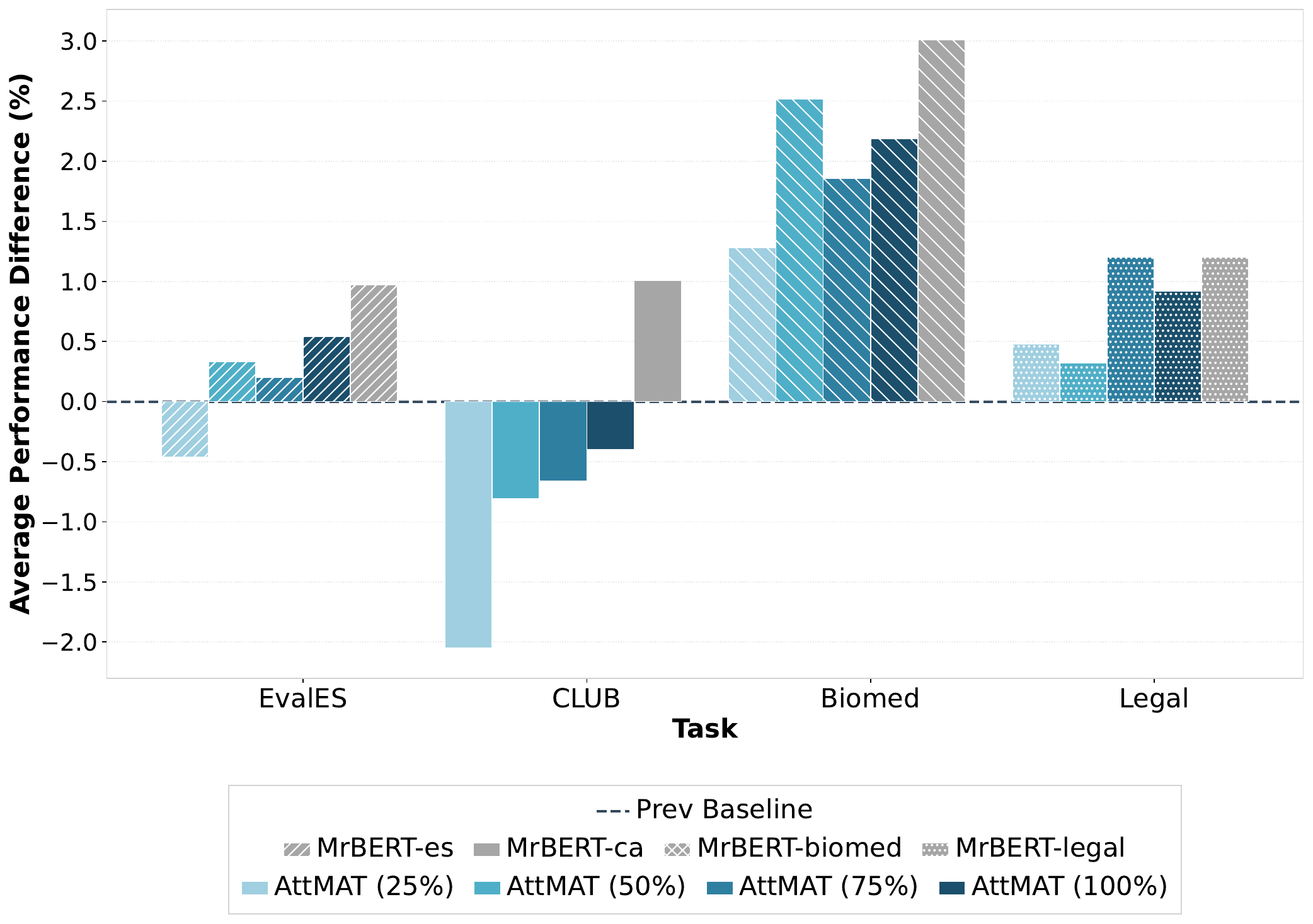}
    \caption{Performance comparison of matryoshka models at different compression levels (25\%, 50\%, 75\%, 100\% of the attention heads) against MrBERT models without matryoshka training across four benchmark tasks. Bars represent the average performance difference to the previous higher benchmark value.}
    \label{fig:matr_all}
\end{figure}

In Figure \ref{fig:matr_all}, we use as baseline the best-performing model in our evaluation that does not belong to the MrBERT family. We then measure the performance gains obtained through domain and vocabulary adaptation, both with and without the matryoshka scheme. For domain adaptation, performance gains are largely preserved under matryoshka, demonstrating the robustness of the adaptation. Notably, the adapted models consistently outperform the baseline even when using only $25\%$ of the attention heads, while achieving up to a $2.4\times$ speedup. 

In contrast, vocabulary adaptation shows weaker resilience to matryoshka compression. This is most severe for Catalan, where MrBERT-ca degrades by 3.06 points at 25\% compression, substantially worse than domain adaptations. Spanish exhibits intermediate degradation: more than domain-adapted models (which retain the original vocabulary) but less than Catalan. We hypothesize this hierarchy reflects a compounding challenge: vocabulary adaptation forces the model to learn new token representations, and matryoshka compression then restricts the representational capacity available for this learning. This dual constraint is particularly punishing for lower-resource languages like Catalan, where limited training data cannot adequately compensate. Our results suggest that for vocabulary-adapted models in lower-resource settings, aggressive compression (25\%) may not justify the performance costs, whereas domain adaptations sustain even heavy pruning with minimal degradation.

\section{Conclusions}

We introduce MrBERT, a family of modern multilingual encoders built on the ModernBERT architecture that achieves robust performance across multilingual, monolingual, and domain-specific evaluations. Through systematic vocabulary adaptation, our compact 150M-parameter Spanish and Catalan models achieve state-of-the-art results (89.83 on EvalES and 85.49 on CLUB) having half of the parameters than the multilingual parent. Our domain-adapted variants for biomedicine and law maintain the full 300M-parameter capacity and consistently outperform existing specialized encoders, demonstrating the effectiveness of continued pre-training on carefully curated domain corpora while preserving broad multilingual capabilities.

Beyond specialization, we integrate Matryoshka representations to address real-world deployment constraints where systems must balance accuracy against latency and storage costs. Our analysis shows that attention-based configurations enable up to 2.4× inference speedup at 25\% capacity while maintaining competitive performance, with domain-adapted models proving more resilient to compression than vocabulary-adapted ones. Ultimately, the MrBERT family demonstrates that modern encoders can simultaneously achieve linguistic excellence, domain expertise, and deployment efficiency, providing practitioners with a principled toolkit for diverse natural language understanding tasks.

\newpage

% \section*{Software and Data}

% If a paper is accepted, we strongly encourage the publication of software and
% data with the camera-ready version of the paper whenever appropriate. This can
% be done by including a URL in the camera-ready copy. However, \textbf{do not}
% include URLs that reveal your institution or identity in your submission for
% review. Instead, provide an anonymous URL or upload the material as
% ``Supplementary Material'' into the OpenReview reviewing system. Note that
% reviewers are not required to look at this material when writing their review.

% Acknowledgements should only appear in the accepted version.
\section*{Acknowledgements}
This project has benefited from the contributions of numerous teams and institutions through data contributions.

In Catalonia, many institutions have been involved in the project. Our thanks to Òmnium Cultural, Parlament de Catalunya, Institut d'Estudis Aranesos, Racó Català, Vilaweb, ACN, Nació Digital, El món and Aquí Berguedà.

At national level, we are especially grateful to our ILENIA project partners: CENID, HiTZ and CiTIUS for their participation. We also extend our genuine gratitude to the Spanish Senate and Congress, Fundación Dialnet, Fundación Elcano, the "Instituto de Ingenieria del Conocimiento" and the ‘Instituto Universitario de Sistemas Inteligentes y Aplicaciones Numéricas en Ingeniería (SIANI)’ of the University of Las Palmas de Gran Canaria.

At the international level, we thank the Welsh government, DFKI, Occiglot project, especially Malte Ostendorff, and The Common Crawl Foundation, especially Pedro Ortiz, for their collaboration.

Finally, we are deeply grateful to the Spanish and Catalan governments for their financial support, which has made this entire endeavor possible. This work has been supported and funded by the Ministerio para la Transformación Digital y de la Función Pública and the Plan de Recuperación, Transformación y Resiliencia – funded by the EU through NextGenerationEU, within the framework of the Modelos del Lenguaje project, it has been promoted and financed by the Government of Catalonia through the Aina project. It is also funded by the Ministerio para la Transformación Digital y de la Función Pública and Plan de Recuperación, Transformación y Resiliencia - Funded by EU – NextGenerationEU within the framework of the project ILENIA with reference 2022/TL22/00215337, 2022/TL22/00215336, 2022/TL22/00215335, 2022/TL22/00215334, as well as by Daniel Tamayo's fellowship within the “Generación D” initiative, Red.es, Ministerio para la Transformación Digital y de la Función Pública, for talent attraction (C005/24-ED CV1). Funded by the European Union NextGenerationEU funds, through PRTR.

\section*{Impact Statement}

This work advances multilingual NLP by developing efficient encoder models for Spanish, Catalan, and specialized domains. We acknowledge the following societal implications:

\paragraph{Positive Impacts.} Our models promote linguistic diversity by providing state-of-the-art performance for mid-resource languages (Spanish and Catalan). The computational efficiency of our language-adapted variants makes advanced language technology more accessible to organizations with limited resources. Domain-adapted models for biomedicine and legal applications may improve information retrieval in high-stakes fields when used appropriately.

\paragraph{Limitations and Risks.} These encoders should not replace expert judgment in medical or legal contexts, they are designed to assist with information retrieval and document organization, not to make clinical or legal decisions. Like all language models trained on web-scale data, they may inherit biases from training corpora, including underrepresentation of dialectal variation (e.g., Latin American Spanish variants) and historical biases in scientific and legal documents. The models could potentially be misused for large-scale document surveillance, biased filtering systems, or retrieval applications that systematically disadvantage certain dialects or writing styles. We recommend human oversight for high-stakes applications and validation for specific use contexts.

\paragraph{Broader Considerations.} While our language-adapted models improve efficiency, the initial pretraining required substantial computational resources. We use only openly licensed data (detailed in Appendix \ref{app:data_sources}) and commit to transparent documentation of model capabilities, limitations, and intended uses to enable responsible deployment.

% In the unusual situation where you want a paper to appear in the
% references without citing it in the main text, use \nocite
% \nocite{langley00}

\bibliography{bibliography}
\bibliographystyle{icml2026}

%%%%%%%%%%%%%%%%%%%%%%%%%%%%%%%%%%%%%%%%%%%%%%%%%%%%%%%%%%%%%%%%%%%%%%%%%%%%%%%
%%%%%%%%%%%%%%%%%%%%%%%%%%%%%%%%%%%%%%%%%%%%%%%%%%%%%%%%%%%%%%%%%%%%%%%%%%%%%%%
% APPENDIX
%%%%%%%%%%%%%%%%%%%%%%%%%%%%%%%%%%%%%%%%%%%%%%%%%%%%%%%%%%%%%%%%%%%%%%%%%%%%%%%
%%%%%%%%%%%%%%%%%%%%%%%%%%%%%%%%%%%%%%%%%%%%%%%%%%%%%%%%%%%%%%%%%%%%%%%%%%%%%%%
\newpage
\appendix
\onecolumn
% \section{You \emph{can} have an appendix here.}

% You can have as much text here as you want. The main body must be at most $8$
% pages long. For the final version, one more page can be added. If you want, you
% can use an appendix like this one.

% The $\mathtt{\backslash onecolumn}$ command above can be kept in place if you
% prefer a one-column appendix, or can be removed if you prefer a two-column
% appendix.  Apart from this possible change, the style (font size, spacing,
% margins, page numbering, etc.) should be kept the same as the main body.
% %%%%%%%%%%%%%%%%%%%%%%%%%%%%%%%%%%%%%%%%%%%%%%%%%%%%%%%%%%%%%%%%%%%%%%%%%%%%%%%
% %%%%%%%%%%%%%%%%%%%%%%%%%%%%%%%%%%%%%%%%%%%%%%%%%%%%%%%%%%%%%%%%%%%%%%%%%%%%%%%

\newpage
\section{Data Sources} \label{app:data_sources}

This appendix presents the datasets used throughout this work for model training at each stage. Full details are given in Table~\ref{tab:datasets}.

\setlength{\tabcolsep}{4pt}        % slightly less horizontal padding
\setlength\LTleft{\fill}
\setlength\LTright{\fill}

\small
\begin{longtable}{@{} 
  %>{\RaggedRight\arraybackslash}
  >{\raggedright\arraybackslash}p{0.23\textwidth}  % name
  %>{\RaggedRight\arraybackslash}
  p{0.09\textwidth}  % languages
  %>{\RaggedRight\arraybackslash}
  p{0.10\textwidth}  % tokens
  %>{\RaggedRight\arraybackslash}
  >{\raggedright\arraybackslash}p{0.19\textwidth}  % usage
  %>{\RaggedRight\arraybackslash}
  p{0.24\textwidth}  % citation (wider)
  %>{\RaggedRight\arraybackslash}
  p{0.08\textwidth}  % url
 @{}} 

\toprule
\textbf{Dataset name} & \textbf{Languages} & \textbf{Domain} & \textbf{Usage} & \textbf{Citation} & \textbf{URL} \\
\midrule
\endfirsthead

\multicolumn{6}{c}%
{\tablename\ \thetable\ -- continued from previous page} \\
\toprule
\textbf{Dataset name} & \textbf{Languages} & \textbf{Domain} & \textbf{Usage} & \textbf{Citation} & \textbf{URL} \\
\midrule
\endhead

\midrule \multicolumn{6}{r}{\textit{continued on next page}} \\
\endfoot

%\bottomrule
\endlastfoot

% Rows

Academic Slovene KAS 2.0 & SL & Education &
Pre-Train
& \cite{erjavec2021kas} & \myurl{https://www.clarin.si/repository/xmlui/handle/11356/1448} \\
\dline

ACAD-Train & All & MT-Science & 
Pre-Train
& \cite{lacunza2025acadata} & \myurl{https://huggingface.co/datasets/BSC-LT/ACAData} \\
\dline

AEPD (juridical resolutions) & ES & Legal & 
Pre-Train
\newline Language Adapatation
\newline Domain Adapatation
& -- & Crawled
\footnote{Crawled from \href{https://www.aepd.es/informes-y-resoluciones/resoluciones}{https://www.aepd.es/informes-y-resoluciones/resoluciones} until 09/2025.} 
\\
\dline

ALIA-TOURISM & ES & General & 
Pre-Train
\newline Language Adapatation
\newline Domain Adapatation
& \cite{alia2025tourism} & \myurl{https://huggingface.co/datasets/gplsi/alia_tourism} \\
\dline
% ALIA\_BOUA & -- & -- & -- & \cite{alia2025boua} & -- \\

% ALIA\_AMIC & -- & -- & -- & \cite{alia2025amic} & -- \\

ALIA-DOGV & ES & General &
Pre-Train
\newline Language Adapatation
\newline Domain Adapatation
 & \cite{alia2025dogv} & \myurl{https://huggingface.co/datasets/gplsi/alia_dogv} \\
\dline

ALIA-Legal-Administrative & ES & Legal & 
Pre-Train
\newline Language Adapatation
\newline Domain Adapatation
 & -- & \href{https://huggingface.co/datasets/SINAI/ALIA-legal-administrative}{URL} \\
\dline

ALIA-LES-CORTS & ES & General &
Pre-Train
\hspace{1.5cm}Language Adapatation
\newline Domain Adapatation
 & \cite{alia2025lescorts} & \myurl{https://huggingface.co/datasets/gplsi/alia_les_corts} \\
\dline

% arXiv (filtered) & EN & Science & 
% Language Adapatation
% \newline Domain Adapatation
% & \cite{kandpal2025common} & \myurl{https://huggingface.co/datasets/common-pile/arxiv_papers_filtered} \\
% \dline

AutoMath & EN & Math & 
Pre-Train
& \cite{zhang2025autonomous} & \myurl{https://huggingface.co/datasets/math-ai/AutoMathText} \\
\dline

Basque Country Official \hspace{1cm}Bulletin  & EU & Legal & 
Pre-Train
& -- & Crawled
\footnote{Crawled from \href{https://www.euskadi.eus/web01-bopv/es/}{https://www.euskadi.eus/web01-bopv/es/} until 09/2025.} 
\\
\dline

Basque Parliament & EU & Legal & 
Pre-Train
& -- & Crawled
\footnote{Crawled from \href{https://www.euskadi.eus/inicio/}{https://www.euskadi.eus/inicio/} until 09/2025.} 
\\
\dline

Berria & EU & General & 
Pre-Train
& -- & Crawled
\footnote{Crawled from \href{https://www.berria.eus/}{https://www.berria.eus/} until 09/2025.} 
\\
\dline

BIGPATENT & EN & General &
Pre-Train
\newline Language Adapatation
& \cite{sharma2019bigpatent} & \myurl{https://huggingface.co/datasets/NortheasternUniversity/big_patent} \\
\dline

Biomed-Enriched  \newline (commercial only) & EN & Biomed & 
Domain Adapatation
& \cite{touchent2025biomed} & \myurl{https://huggingface.co/datasets/almanach/Biomed-Enriched} \\
\dline

Booktegi & EU & Books & 
Pre-Train
& -- & Crawled
\footnote{Crawled from \href{https://www.booktegi.eus/}{https://www.booktegi.eus/} until 09/2025.} 
\\
\dline

Brazilian Portuguese Web as Corpus (BrWaC) & PT & General &
Pre-Train
& \cite{wagner-filho-etal-2018-brwac} & \myurl{https://www.inf.ufrgs.br/pln/wiki/index.php?title=BrWaC} \\
\dline

Bulgarian National Corpus (BulNC) & BG & General &
Pre-Train
& -- & \myurl{http://old.dcl.bas.bg/dataset/BulNC.7z} \\
\dline

CaBeRnet & FR & General &
Pre-Train
& \cite{popa-fabre-etal-2020-french} & -- \\
\dline

CATalog 1.0 & CA & General &
Pre-Train
& \cite{palomar2024curated} & \myurl{https://huggingface.co/datasets/projecte-aina/CATalog} \\
\dline

CARMEN-I & ES & Biomed & 
Language Adapatation
\newline Domain Adapatation
& \cite{farre_maduell_2024_carmen_i} & \myurl{https://physionet.org/content/carmen-i/1.0.1/} \\
\dline

Colossal OSCAR
\footnote{06-07-22 \& 05-06-23 chunks.} - Basque 
& EU & General & 
Pre-Train
& \cite{brack2024communityoscar} & \myurl{https://huggingface.co/datasets/oscar-corpus/community-oscar} \\

&  &  & 

&  &  \\

CorpusNÓS & GL & General &
Pre-Train
& \cite{de2024corpusnos} & -- \\
\dline

CoQCat & CA & QA &
Pre-Train
& \cite{gonzalez2024building} & \myurl{https://huggingface.co/datasets/projecte-aina/CoQCat} \\
\dline

Croatian Web as Corpus 2.1 (hrWaC) & CR & General &
Pre-Train
& \cite{ljubevsic2011hrwac} & \myurl{https://nlp.ffzg.unizg.hr/resources/corpora/hrwac/} \\
\dline

CulturaX & EU & Culture & 
Pre-Train
& \cite{nguyen-etal-2024-culturax} & \myurl{https://huggingface.co/datasets/uonlp/CulturaX} \\
\dline

CURLICAT & BG,CR,HU, PL, RO, SL, SK & General &
Pre-Train
& \cite{varadi2022introducing} & \myurl{https://curlicat-project.eu/} \\
\dline

C4 (Basque only) & EU & General & 
Pre-Train
& -- & \myurl{https://huggingface.co/datasets/allenai/c4} \\
\dline

DaNewsroom & DA & General &
Pre-Train
& \cite{varab2020danewsroom} & \myurl{https://github.com/danielvarab/da-newsroom} \\
\dline

Danish GigaWord & DA & General &
Pre-Train
& \cite{derczynski2021danish} & \myurl{https://sprogteknologi.dk/dataset/danish-gigaword} \\
\dline

DK-CLARIN Reference \hspace{1cm}Corpus of General Danish & DA & General &
Pre-Train
& -- & \myurl{https://korpus.dsl.dk/clarin/} \\
\dline

Egunkaria & EU & General & 
Pre-Train
& -- & Crawled
\footnote{Content from the daily Basque newspaper Euskaldunon Egunkaria (2001–2006).} 
\\
\dline

Estonian National Corpus 2021 (ENC) & ET & General &
Pre-Train
& \cite{koppel2019skell} & \myurl{https://metashare.ut.ee/repository/browse/estonian-national-corpus-2021-vert/4547c7bea0d411eebb4773db10791bcfd961b8c70b544966800142b04f957a86/} \\
\dline

Estonian Reference Corpus (ERC) & ET & General &
Pre-Train
& -- & \myurl{https://www.cl.ut.ee/korpused/segakorpus/} \\
\dline

EURLEX-Resources & All & Legal & 
Pre-Train
\newline Language Adapatation
\newline Domain Adapatation
& -- & \myurl{https://huggingface.co/datasets/joelniklaus/eurlex_resources} \\
\dline

Europarl & All & MT-Legal & 
Pre-Train
& \cite{tiedemann-2012-parallel} & \myurl{https://huggingface.co/datasets/Helsinki-NLP/europarl} \\
\dline

EusCrawl (w/o Wikipedia or NC-licenses) & EU & General & 
Pre-Train
& \cite{artetxe2022euscrawl} & \myurl{https://huggingface.co/datasets/HiTZ/euscrawl} \\
\dline

FineMath-4+ & EN & Math & 
Pre-Train
& \cite{allal2025smollm2smolgoesbig} & \myurl{https://huggingface.co/datasets/HuggingFaceTB/finemath} \\
\dline

FinePDFs - Basque & EU & General & 
Pre-Train
& \cite{kydlicek2025finepdfs} & \myurl{https://huggingface.co/datasets/HuggingFaceFW/finepdfs} \\
\dline

FineWeb-EDU \hspace{1.5cm}(highest-quality documents) & EN & Education &
Pre-Train
\newline Language Adapatation
& \cite{lozhkov2024fineweb-edu} & \myurl{https://huggingface.co/datasets/HuggingFaceFW/fineweb-edu-score-2} \\
\dline

FineWeb2 & All & General & 
Pre-Train
\newline Language Adapatation
& \cite{penedo2025fineweb2pipelinescale} & \myurl{https://huggingface.co/datasets/HuggingFaceFW/fineweb-2} \\
\dline

FineWeb2-HQ & DA, DE, EL, ES, FR, HU, IT, NL, PL, PT, RU, SV & General & 
Pre-Train
\newline Language Adapatation
& \cite{messmer2025multilingdatacomp} & \myurl{https://huggingface.co/datasets/epfml/FineWeb2-HQ} \\
\dline

French Public Domain Books (French-PD) & FR & General &
Pre-Train
& -- & \myurl{https://huggingface.co/datasets/PleIAs/French-PD-Books} \\
\dline

French Public Domain \hspace{1cm}Newspapers (French-PD) & FR & General &
Pre-Train
& -- & \myurl{https://huggingface.co/datasets/PleIAs/French-PD-Newspapers} \\
\dline
% &  & & &  &  \\

German Web as Corpus (DeWaC) & DE & General &
Pre-Train
& -- & \myurl{https://docs.sslmit.unibo.it/doku.php?id=corpora:dewac} \\
\dline

Gipuzkoa Provincial Council  & EU & Legal & 
Pre-Train
& -- & Crawled
\footnote{Crawled from \href{https://egoitza.gipuzkoa.eus/web/council}{https://egoitza.gipuzkoa.eus/web/council} until 09/2025.} 
\\
\dline

Greek Legal Code (GLC) & EL & Legal &
Pre-Train
& \cite{papaloukas2021multi} & -- \\

 &  &  &

&  &  \\

Greek Web Corpus (GWC) & EL & General &
Pre-Train
& \cite{outsios2018word} & \myurl{http://nlp.polytechnique.fr/resources-greek} \\
\dline

HPLT v1 \& v2 - Basque & EU & General & 
Pre-Train
& \cite{de2024new} & v1:\myurl{https://hplt-project.org/datasets/v1} v2:\myurl{https://hplt-project.org/datasets/v2.0}\\
\dline
HPLT v1 - Spanish & ES & General & 
Pre-Train
\newline Language Adapatation
& \cite{de2024new} & \myurl{https://hplt-project.org/datasets/v1} \\
\dline
HPLT v1.1 - Spanish & ES & General & 
Pre-Train
\newline Language Adapatation
& \cite{de2024new} & \myurl{https://hplt-project.org/datasets/v1}\\
\dline

Institutional books \newline (legal \& biomedical) & EN, ES & Legal Biomed & 
Pre-Train
\newline Language Adapatation
\newline Domain Adapatation
& \cite{cargnelutti2025institutionalbooks10242b} & \myurl{https://huggingface.co/datasets/institutional/institutional-books-1.0} \\
\dline

Irish Universal Dependencies (Ga-UD) & GA & General &
Pre-Train
& -- & \myurl{https://universaldependencies.org/ga/index.html} \\
\dline

Italian Web as Corpus (ItWaC) & IT & General &
Pre-Train
& -- & \myurl{https://docs.sslmit.unibo.it/doku.php?id=corpora:itwac} \\
\dline

Korpus Malti & MT & General &
Pre-Train
& \cite{micallef2022pre} & \myurl{https://huggingface.co/datasets/MLRS/korpus_malti} \\
\dline

Korpus slovenských právnych predpisov v1.9 (SK-Laws) & SK & Legal &
Pre-Train
& -- & \myurl{https://www.juls.savba.sk/data.html} \\
\dline

Laws and legal acts of Ukraine (UK-Laws) & UK & Legal &
Pre-Train
& -- & \myurl{https://lang.org.ua/en/corpora/} \\
\dline

LegalACT & ES & Legal & 
Domain Adapatation
& -- & \myurl{https://legal.ds.inesdata-project.eu/catalog/conn-iic/LegalACT Prueba} \\
\dline
MaCoCu & All & General &
Pre-Train
& \cite{banon2022macocu} & \myurl{https://macocu.eu/} \\
\dline

Math AMPS & EN & Math & 
Pre-Train
\newline Language Adapatation
& \cite{hendrycksmath2021} & \myurl{https://github.com/hendrycks/math} \\
\dline

MathPile (Commercial) & EN & Math & 
Pre-Train
& \cite{wang2024mathpile} & \myurl{https://huggingface.co/datasets/GAIR/MathPile_Commercial} \\
\dline

MARCELL Romanian \hspace{1cm}legislative subcorpus v2 & RO & Legal &
Pre-Train
& -- & \myurl{https://elrc-share.eu/repository/browse/marcell-romanian-legislative-subcorpus-v2/2da548428b9d11eb9c1a00155d026706ce94a6b59ffc4b0e9fb5cd9cebe6889e/} \\
\dline

MedlinePlus & EN & Biomed & 
Domain Adapatation
& -- & Crawled
\footnote{Crawled from \href{https://medlineplus.gov/}{https://medlineplus.gov/} until 09/2025.} 
\\
\dline

MC4 Legal & EN & Legal & 
Pre-Train
\newline Language Adapatation
& -- & \href{https://huggingface.co/datasets/joelniklaus/legal-mc4}{URL} \\
\dline

News Commentary & All & MT-General & 
Pre-Train
& \cite{kocmi-etal-2022-findings} & \myurl{https://huggingface.co/datasets/wmt/news-commentary} \\
\dline

NKPJ National Corpus of \hspace{1cm}Polish v1.2 (NKPJ) & PL & General &
Pre-Train
& \cite{lewandowska2013national} & \myurl{https://clip.ipipan.waw.pl/NationalCorpusOfPolish} \\
\dline

Norwegian Colossal Corpus (NCC) & NO & General &
Pre-Train
& \cite{kummervold2021operationalizing} & \myurl{https://huggingface.co/datasets/NbAiLab/NCC} \\
\dline

Occitan Corpus (IEA-AALO) & OC & General &
Pre-Train
& -- & -- \\
\dline

Official Gazette of the \hspace{1cm}Historical Territory of Alava & EU & Legal & 
Pre-Train
& -- & Crawled
\footnote{Crawled from \href{https://www.araba.eus/botha/inicio/sgbo5001.aspx}{https://www.araba.eus/botha/inicio/sgbo5001.aspx} until 09/2025.} 
\\
\dline

OpenSubtitles v2016 & EN & General & 
Pre-Train
\newline Language Adapatation
& \cite{lison-tiedemann-2016-opensubtitles2016} & \myurl{http://opus.lingfil.uu.se/OpenSubtitles2016.php} \\
\dline

OpenSubs v2018 - Basque & EU & General & 
Pre-Train
& -- & \myurl{https://opus.nlpl.eu/legacy/OpenSubtitles-v2018.php} \\
\dline

OpenWeb (math subset) & EN & Math & 
Pre-Train
& \cite{paster2023openwebmath} & \myurl{https://huggingface.co/datasets/open-web-math/open-web-math} \\
\dline

Open Legal Data - German court decisions and laws & DE & Legal &
Pre-Train
& \cite{ostendorff2020towards} & \myurl{https://openlegaldata.io/} \\

 & & & & & \\

ParlamentoPT & PT & Legal &
Pre-Train
& \cite{rodrigues2023advancing} & \myurl{https://huggingface.co/datasets/PORTULAN/parlamento-pt} \\
\dline

Parlamint & All & Legal &
Pre-Train
& \cite{erjavec2023parlamint} & \myurl{https://www.clarin.eu/parlamint} \\
\dline

% pes2o & EN & Science & 
% Domain Adapatation
% & \cite{peS2o} & \myurl{https://huggingface.co/datasets/allenai/peS2o} \\
% \dline

PG-19 & EN & Books & 
Pre-Train
\newline Language Adapatation
& \cite{rae2019compressive} & \myurl{https://huggingface.co/datasets/deepmind/pg19} \\
\dline

Pile of Law & EN & Legal & 
Pre-Train
\newline Language Adapatation
& \cite{henderson2022pile} & \myurl{https://huggingface.co/datasets/pile-of-law/pile-of-law} \\
\dline

Polish Parliamentary Corpus (PPC) & PL & Legal &
Pre-Train
& \cite{ogrodniczuk2018polish} & \myurl{https://clip.ipipan.waw.pl/PPC} \\
\dline

Proof Pile & EN & Math & 
Pre-Train
\newline Language Adapatation
& -- & \href{https://huggingface.co/datasets/hoskinson-center/proof-pile}{URL} \\
\dline

PubMed (abstracts) - Spanish & ES & Biomed & 
Pre-Train
\newline Language Adapatation
\newline Domain Adapatation
& -- & Crawled
\footnote{Crawled from \href{https://pubmed.ncbi.nlm.nih.gov/}{https://pubmed.ncbi.nlm.nih.gov/} until 09/2025.} 
\\
\dline

Recolecta (train) & EN, ES &  Legal Biomed & 
Domain Adapatation
& -- & Crawled
\footnote{Full explanation on Appendix~\ref{app:recolecta}.} 
\\

\dline

SK Court Decisions v2.0 \hspace{1cm}(OD-Justice) & SK & Legal &
Pre-Train
& -- & \myurl{https://www.juls.savba.sk/data/od-justice/} \\
\dline

Slovene Web as Corpus (slWaC) & SL & General &
Pre-Train
& \cite{erjavec2015slwac} & \myurl{https://nlp.ffzg.unizg.hr/resources/corpora/slwac/} \\
\dline

SoNaR Corpus NC 1.2 & NL & General &
Pre-Train
& -- & \myurl{https://taalmaterialen.ivdnt.org/download/tstc-sonar-nieuwe-media-corpus-1/} \\ \dline

Spanish-Legal-Data-2 & ES & Legal & 
Pre-Train
\newline Language Adapatation
\newline Domain Adapatation
& \cite{ramitha_spanish_legal_data_2} & \href{https://huggingface.co/datasets/Ramitha/spanish-legal-data-2}{URL} \\
\dline

Spanish Legal Domain Corpora & ES & Legal &
Pre-Train
\newline Language Adapatation
\newline Domain Adapatation
& -- & \myurl{https://github.com/PlanTL-GOB-ES/lm-legal-es} \\
\dline

SrpKorSubset: news, legal, \hspace{1cm}academic, conversation, literary
(SrpKor) & SR & General &
Pre-Train
& -- & \myurl{http://www.korpus.matf.bg.ac.rs} \\
\dline

State-related content from the Latvian Web (State-Latvian-
Web) & LT & Legal &
Pre-Train
& -- & \myurl{http://catalog.elra.info/en-us/repository/browse/ELRA-W0169/} \\
\dline

SYN v9: large corpus of written Czech & CZ & General &
Pre-Train
& \cite{11234/1-4635} & \myurl{lindat.mff.cuni.cz/repository/xmlui/handle/11234/1-4635} \\
\dline

Tagesschau Archive Article & DE & General &
Pre-Train
& -- & \myurl{https://huggingface.co/datasets/bjoernp/tagesschau-2018-2023} \\
\dline

The Danish Parliament Corpus 2009 - 2017, v1 & DA & -- &
Pre-Train
& \cite{hansen2018danish} & \myurl{https://repository.clarin.dk/repository/xmlui/handle/20.500.12115/8} \\
\dline
%  & & &
% &  &  \\

StarCoder & Code & Code & 
Pre-Train
& \cite{li2023starcoder} & \myurl{https://huggingface.co/datasets/bigcode/the-stack-dedup} \\
\dline

The Gaois bilingual corpus of English-Irish legislation \hspace{1cm}(Ga-Legislation) & EN, GA & Legal &
Pre-Train
& -- & \myurl{https://portulanclarin.net/repository/browse/the-gaois-bilingual-corpus-of-english-irish-legislation-processed/daeac17c9e3511ea9b7f02420a000407b83de243dc0b469aab41084386c5b80f/} \\
\dline

The Pile (PhilPapers subset) & EN & Education & 
Pre-Train
\newline Language Adapatation
& \cite{gao2020pile} & \myurl{https://github.com/thoppe/The-Pile-PhilPapers} \\

 & & & & & \\

The Swedish Culturomics \hspace{1cm}Gigaword Corpus (Swedish-
Gigaword) & SW & General &
Pre-Train
& \cite{eide2016swedish} & \myurl{https://spraakbanken.gu.se/en/resources/gigaword} \\
\dline

Welsh-GOV & CY & Legal &
Pre-Train
& -- & \myurl{https://www.llyw.cymru/} \\
\dline

Wikimedia dumps & All & General & 
Pre-Train
\newline Language Adapatation
& -- & \href{https://dumps.wikimedia.org/}{URL} \\
\dline

Yle Finnish News Archive \hspace{1cm}(Yle-News) & FI & General &
Pre-Train
& -- & \myurl{http://urn.fi/urn:nbn:fi:lb-2021050401} \\
\dline

Zelai Handi & EU & General & 
Pre-Train
& \cite{ZelaiHandi} & \myurl{https://huggingface.co/datasets/orai-nlp/ZelaiHandi} \\
\dline

3CEL & ES & Legal & 
Domain Adapatation
& \cite{garcia20253cel} & \myurl{https://legal.ds.inesdata-project.eu/catalog/conn-iic/3CEL Corpus} \\

\bottomrule
\caption{Data sources used throughout this work.}\label{tab:datasets}
\end{longtable}

\section{Language Distribution} \label{app:lang_sources}
\begin{table}[H]
\centering
\begin{tabular}{cccccc}
\hline
\textbf{Language} & \textbf{Tokens}   & \textbf{Language} & \textbf{Tokens} & \textbf{Language} & \textbf{Tokens}   \\ \hline
EN                & 4,120,759,329,876 & PL                & 44,252,717,879  & LT                & 7,017,976,396     \\
ES                & 523,633,854,143   & CS                & 37,284,432,759  & EU                & 6,999,041,964     \\
DE                & 174,456,549,084   & BG                & 33,060,589,015  & NO                & 6,798,808,558     \\
FR                & 172,729,856,318   & CA                & 26,664,618,307  & GL                & 5,173,500,585     \\
CODE              & 156,565,808,119   & RO                & 24,395,563,081  & LV                & 4,970,822,927     \\
RU                & 141,760,491,684   & SK                & 21,968,084,510  & TRANSLATIONS      & 3,800,022,519     \\
HU                & 121,749,464,875   & SV                & 20,634,419,489  & MT                & 1,627,139,688     \\
MATH              & 85,020,827,274    & FI                & 16,516,320,096  & CY                & 945,882,400       \\
IT                & 73,270,944,125    & DA                & 12,673,977,209  & GA                & 638,247,061       \\
PT                & 72,051,734,796    & SL                & 10,415,839,612  & SH                & 395,040,116       \\
UK                & 57,081,052,350    & SR                & 9,936,144,706   & NN                & 214,056,022       \\
EL                & 53,452,523,986    & ET                & 7,820,108,307   & OC                & 191,488,793       \\
NL                & 46,330,125,515    & HR                & 7,228,374,303   & Total             & 6,110,485,778,447 \\
\hline
\end{tabular}
\caption{Token distribution by language during Pre-Training phase.}
\end{table}

\newpage
\section{Classifier Categories} \label{app:classifier_categories}
Table~\ref{tab:domain_classifier_description} provides descriptions of the domains predicted by NVIDIA’s Multilingual Domain Classifier

\begin{table*}[ht]
\centering

\begin{tabular}{p{4.2cm}p{10.5cm}}
\toprule
\textbf{Domain Class} & \textbf{Description} \\
\midrule
Adult & Sexual content, pornography, or age-restricted material \\
Arts\_and\_Entertainment & Music, movies, theater, celebrities, pop culture \\
Autos\_and\_Vehicles & Cars, motorbikes, vehicle news and reviews \\
Beauty\_and\_Fitness & Skincare, cosmetics, wellness, workout routines \\
Books\_and\_Literature & Novels, literary criticism, poetry, book reviews \\
Business\_and\_Industrial & Enterprise, corporate, manufacturing, B2B topics \\
Computers\_and\_Electronics & Hardware, software, tech news, consumer gadgets \\
Finance & Banking, investing, personal finance, stock markets \\
Food\_and\_Drink & Recipes, restaurants, food culture, drinks \\
Games & Video games, board games, eSports, gaming culture \\
Health & Medical topics, mental health, wellness, diseases \\
Hobbies\_and\_Leisure & DIY, crafts, hobbies, leisure activities \\
Home\_and\_Garden & Home improvement, gardening, decor \\
Internet\_and\_Telecom & ISPs, web platforms, telecommunications \\
Jobs\_and\_Education & Career guidance, job listings, academic topics \\
Law\_and\_Government & Legislation, public policy, political topics \\
News & Journalism, current events, news reporting \\
Online\_Communities & Forums, social platforms, user communities \\
People\_and\_Society & Culture, social issues, demographics \\
Pets\_and\_Animals & Pet care, wildlife, zoology topics \\
Real\_Estate & Property listings, housing market, realty advice \\
Science & Research, scientific articles, STEM topics \\
Sensitive\_Subjects & Controversial or delicate content (e.g. abuse, violence) \\
Shopping & E-commerce, product reviews, retail \\
Sports & Athletic events, scores, sports commentary \\
Travel\_and\_Transportation & Tourism, transit, travel guides \\
\bottomrule
\end{tabular}
\caption{Domain class descriptions for NVIDIA’s multilingual domain classifier, based on manual inspection of sample instances.}
\label{tab:domain_classifier_description}

\end{table*}

\newpage
\section{Hyperparameters Settings}\label{app:hyper}
\begin{table*}[h]
\begin{tabular}{cccccc}
\hline
                     & MrBERT            & MrBERT-es         & MrBERT-ca       & MrBERT-biomed   & MrBERT-legal    \\ \hline
Scheduler            & WSD               & WSD               & Warmup + Cosine & Warmup + Cosine & Warmup + Cosine \\
Learning Rate        & 1e-3              & 4e-4              & 1e-3            & 2e-3            & 3e-3            \\
Total Tokens         & 6,100B            & 615B              & 47.4B           & 24.1B           & 9B              \\
Epochs               & 1                 & 1                 & 1               & 2               & 10              \\
Warmup Tokens        & 3B                & 3B                & 4.7B            & 2.4B            & 9B              \\
Decay Tokens         & 100B              & 100B              & 42.7B           & 45.9B           & 81B             \\
Number of Parameters & 308M              & 150M              & 150M            & 308M            & 308M            \\
MLM Probability      & 0.3 (WS), 0.1 (D) & 0.3 (WS), 0.1 (D) & 0.1             & 0.1             & 0.3             \\
Samples/s (25\% heads)  & 34.2              & 47.0              & 47.0            & 34.2            & 34.2            \\
Samples/s (50\% heads)  & 23.4              & 28.3              & 28.3            & 23.4            & 23.4            \\
Samples/s (75\% heads)  & 17.7              & 20.4              & 20.4            & 17.7            & 17.7            \\
Samples/s (100\% heads) & 14.2              & 15.9              & 15.9            & 14.2            & 14.2            \\ \hline
\end{tabular}
\caption{List of hyperparameters chosen for each model. Throughput measurements (samples/s) were obtained from speed tests launched on a single NVIDIA H100 GPU with 64\,GB of memory. Reported values have an estimated error bar of $\pm 0.1$ samples/s. Each inference sample consists of 8{,}192 tokens.}
\end{table*} 

% \newpage
\section{Recolecta} \label{app:recolecta}

This appendix introduces how the training and evaluation dataset named as ``Recolecta'' throughout this work was obtained and how it was afterwards divided into a train and test split.

\subsection{Dataset Creation}

The national aggregator of open-access scientific repositories, RECOLECTA\footnote{\href{https://recolecta.fecyt.es/}{https://recolecta.fecyt.es/}}, was used to source scientific documents in PDF format. Documents were crawled through April 2025, and the corpus was restricted to publications in English and Spanish.

To ensure legal compliance for data processing and redistribution, the documents were filtered based on their license metadata. Only documents with licenses permitting both publication and training, or at least training, were included:

\begin{itemize}
    \item \textbf{Licenses permitted for Publishing and Training}: CC-BY, CC0, Apache, BSD, MIT, and Open Access.
    \item \textbf{Licenses permitted for Training}: GPL, SA (ShareAlike), and documents with specific Catalan permissions allowing reproduction and communication for derivative works.
\end{itemize}

Documents marked with restrictive terms such as ``NoDerivatives'' (ND), ``Non-Commercial'' (NC), ``All Rights Reserved,'' or ``Restricted Access'' were excluded from the final dataset.

This filtering process resulted in a total of 673,814 PDFs. To extract the textual content from these documents while maintaining structural integrity, the olmOCR model\footnote{\href{https://huggingface.co/allenai/olmOCR-7B-0725}{https://huggingface.co/allenai/olmOCR-7B-0725}} was employed.

\subsection{AbSanitas subset}

From the filtered RECOLECTA corpus, the AbSanitas dataset (introduced in Section 5.1) was constructed by excluding non-scientific repositories and extracting biomedical abstracts from these sources. This selection was performed mainly using document metadata and abstract-level semantic cues to ensure the resulting subset was restricted to the biomedical domain.

The resulting dataset contains 12{,}596 documents, each associated with two queries. Documents were split at the document level into train (10{,}076), development (1{,}259), and test (1{,}261) sets, corresponding to an 80/10/10 split, ensuring the absence of document overlap across sets. Splits were also created prior to query generation to prevent leakage.

\newpage
\section{Matryoshka Results} \label{app:matry}

\begin{table}[h]
\centering
\begin{tabular}{lcccccccc}
\hline
\textbf{Model} & \textbf{UD-POS} & \textbf{XNLI} & \textbf{PAWS-X} & \textbf{PANX} & \textbf{TyDiQA} & \textbf{MLQA} & \textbf{XQuAD} & \textbf{Average} \\
 & \textbf{(F1)} & \textbf{(Acc.)} & \textbf{(Acc.)} & \textbf{(F1)} & \textbf{(F1)} & \textbf{(F1)} & \textbf{(F1)} & \\
\hline
MrBERT & \underline{83.74} & \underline{81.26} & 91.32 & 72.06 & \textbf{56.34} & \textbf{70.67} & 77.91 & 76.19 \\
MrBERT AttMAT (25\%) & 81.79 & 79.57 & 89.85 & 68.51 & 46.79 & 67.10 & 72.99 & 72.37 \\
MrBERT MAT (25\%) & 80.53 & 79.08 & 88.98 & 68.27 & 49.05 & 66.98 & 73.73 & 72.37 \\
MrBERT AttMAT (50\%) & \textbf{87.77} & 80.92 & 91.53 & 70.46 & 53.49 & 69.04 & 76.24 & 75.63 \\
MrBERT MAT (50\%) & 81.71 & 80.31 & 90.38 & 70.47 & 53.63 & 68.70 & 76.48 & 74.53 \\
MrBERT AttMAT (75\%) & 82.23 & 80.86 & 91.28 & 71.61 & 53.07 & 69.82 & 76.95 & 75.12 \\
MrBERT MAT (75\%) & 83.12 & 81.21 & 90.88 & 71.67 & 55.60 & 69.91 & 76.98 & 75.62 \\
MrBERT AttMAT (100\%) & 82.66 & 80.89 & \textbf{92.11} & \underline{72.06} & 54.78 & \underline{70.49} & \textbf{78.06} & \underline{75.87} \\
MrBERT MAT (100\%) & 82.96 & \textbf{81.94} & \underline{91.61} & \textbf{73.61} & \underline{55.93} & 70.09 & \underline{77.96} & \textbf{76.30} \\
\hline
\end{tabular}
\caption{Evaluation results on the Xtreme benchmark over different experiments using Matryoshka in MrBERT.}
\label{tab:your_label}
\end{table}

\begin{table*}[h]
\centering
\small
\begin{tabular}{lcccccccc}
\hline
\textbf{Model} & \textbf{\begin{tabular}[c]{@{}c@{}}AnCora-ca\\-ner (F1)\end{tabular}} & \textbf{\begin{tabular}[c]{@{}c@{}}AnCora-ca\\-pos (F1)\end{tabular}} & \textbf{\begin{tabular}[c]{@{}c@{}}STS-ca\\(Pearson)\end{tabular}} & \textbf{\begin{tabular}[c]{@{}c@{}}TeCla\\(Acc.)\end{tabular}} & \textbf{\begin{tabular}[c]{@{}c@{}}TECA\\(Acc.)\end{tabular}} & \textbf{\begin{tabular}[c]{@{}c@{}}ViquiQuAD\\(F1)\end{tabular}} & \textbf{\begin{tabular}[c]{@{}c@{}}XQuAD\\(F1)\end{tabular}} & \textbf{Average} \\ \hline
MrBERT & {\ul 87.32} & \textbf{99.01} & 83.00 & 73.79 & \textbf{84.03} & \textbf{89.25} & \textbf{73.96} & \textbf{84.34} \\
MrBERT AttMAT (25\%) & 86.09 & 98.87 & 77.69 & 74.02 & 79.55 & 85.82 & 68.60 & 81.52 \\
MrBERT MAT (25\%) & 85.07 & 98.75 & 79.31 & 72.47 & 78.22 & 86.41 & 69.55 & 81.40 \\
MrBERT AttMAT (50\%) & 86.71 & 98.87 & 78.81 & \textbf{74.13} & {\ul 83.66} & 87.49 & 71.11 & 82.97 \\
MrBERT MAT (50\%) & \textbf{87.40} & 98.86 & 82.16 & 73.23 & 81.62 & 88.56 & 70.03 & 83.12 \\
MrBERT AttMAT (75\%) & 87.20 & 98.86 & 81.73 & {\ul 74.03} & 82.48 & 87.87 & 72.57 & 83.53 \\
MrBERT MAT (75\%) & 86.67 & 98.91 & {\ul 83.16} & 73.20 & 82.33 & 88.89 & 72.79 & 83.71 \\
MrBERT AttMAT (100\%) & 86.80 & 98.92 & 82.73 & 73.73 & 82.14 & 88.50 & {\ul 73.82} & 83.81 \\
MrBERT MAT (100\%) & 86.92 & {\ul 98.92} & \textbf{83.19} & 73.63 & 82.76 & {\ul 89.15} & 73.74 & {\ul 84.05} \\ \hline
MrBERT-ca & 88.04 & \textbf{99.03} & \textbf{85.42} & \textbf{74.97} & \textbf{86.92} & \textbf{89.59} & \textbf{74.47} & \textbf{85.49} \\
MrBERT-ca AttMAT (25\%) & 87.34 & 98.94 & 79.67 & 74.29 & 80.49 & 86.83 & 69.45 & 82.43 \\
MrBERT-ca AttMAT (50\%) & {\ul 88.17} & 98.96 & 81.23 & {\ul 74.71} & 81.62 & 88.60 & 72.41 & 83.67 \\
MrBERT-ca AttMAT (75\%) & \textbf{88.32} & 98.93 & 81.06 & 74.31 & 83.23 & 88.38 & 72.49 & 83.82 \\
MrBERT-ca AttMAT (100\%) & 86.98 & {\ul 99.01} & {\ul 82.82} & 74.52 & {\ul 83.51} & {\ul 88.62} & {\ul 73.13} & {\ul 84.08} \\ \hline
\end{tabular}
\caption{Evaluation results on the CLUB benchmark over different experiments using Matryoshka in MrBERT and MrBERT-ca.}
\label{tab:combined-club-results}
\end{table*}

\begin{table}[h!]
\begin{tabular}{lcccccccc}
\hline
\textbf{Model} & \textbf{\begin{tabular}[c]{@{}c@{}}UD-POS\\-es (F1)\end{tabular}} & \textbf{\begin{tabular}[c]{@{}c@{}}CoNLL\\-NERC-es\\(F1)\end{tabular}} & \textbf{\begin{tabular}[c]{@{}c@{}}STS-es\\(Pearson)\end{tabular}} & \textbf{\begin{tabular}[c]{@{}c@{}}PAWS-X\\-es (Acc.)\end{tabular}} & \textbf{\begin{tabular}[c]{@{}c@{}}MlDoc\\(Acc.)\end{tabular}} & \textbf{\begin{tabular}[c]{@{}c@{}}Massive\\(Acc.)\end{tabular}} & \textbf{\begin{tabular}[c]{@{}c@{}}SQAC\\(F1)\end{tabular}} & \textbf{Average} \\ \hline
MrBERT & \textbf{99.06} & 87.42 & \underline{84.18} & 91.25 & 95.28 & \underline{87.46} & \textbf{81.96} & \textbf{89.52} \\
MrBERT AttMAT (25\%) & 99.02 & 86.8 & 78.60 & 89.70 & 96.05 & 86.31 & 76.24 & 87.53 \\
MrBERT MAT (25\%) & 98.95 & 86.29 & \textbf{84.64} & 90.30 & 94.65 & 86.42 & 77.79 & 88.43 \\
MrBERT AttMAT (50\%) & 99.04 & 86.44 & 82.44 & 90.75 & 96.07 & 87.09 & 78.49 & 88.62 \\
MrBERT MAT (50\%) & 99.01 & 86.26 & 82.28 & 90.65 & 95.6 & 86.82 & 78.56 & 88.46 \\
MrBERT AttMAT (75\%) & 99.02 & \underline{87.45} & 83.03 & 90.40 & 95.97 & 86.68 & 79.74 & 88.90 \\
MrBERT MAT (75\%) & 99.02 & 87.29 & 83.69 & 91.40 & \underline{96.25} & 87.39 & 80.26 & \underline{89.33} \\
MrBERT AttMAT (100\%) & 99.03 & \textbf{87.63} & 82.46 & \textbf{91.60} & \textbf{96.28} & 86.95 & 80.45 & 89.20 \\
MrBERT MAT (100\%) & \underline{99.06} & 87.27 & 83.17 & \underline{91.45} & 96.13 & \textbf{87.49} & \underline{80.54} & 89.30 \\ \hline
MrBERT-es & \textbf{99.08} & \textbf{87.77} & \textbf{85.23} & \underline{91.90} & 95.55 & 87.05 & \textbf{82.19} & \textbf{89.83} \\
MrBERT-es AttMAT (25\%) & \underline{99.05} & 87.07 & 82.03 & 89.90 & 95.95 & \underline{87.16} & 77.65 & 88.40 \\
MrBERT-es AttMAT (50\%) & 99.03 & 87.43 & \underline{84.64} & 91.05 & 95.30 & \textbf{87.22} & 79.69 & 89.19 \\
MrBERT-es AttMAT (75\%) & 99.04 & 87.51 & 81.57 & 91.20 & \underline{96.15} & 86.31 & 81.63 & 89.06 \\
MrBERT-es AttMAT (100\%) & 99.02 & \underline{87.58} & 82.09 & \textbf{91.95} & \textbf{96.17} & 87.02 & \underline{81.94} & \underline{89.40} \\ \hline
\end{tabular}
\caption{Evaluation results on the EvalES benchmark over different experiments using Matryoshka in MrBERT and MrBERT-es.}
\end{table}

\begin{table}[h]
\begin{tabular}{
ccccccccccc }
\\ \hline
                                     & \begin{tabular}[c]{@{}c@{}}bsc-bio-\\distemist\\-ner\\(ES)\end{tabular} & \begin{tabular}[c]{@{}c@{}}cantemist\\(ES)\end{tabular} & \begin{tabular}[c]{@{}c@{}}pharma\\-coner\\(ES)\end{tabular} & \begin{tabular}[c]{@{}c@{}}AbSanitas\\(ES)\end{tabular} & \begin{tabular}[c]{@{}c@{}}R2Med\\(EN)\end{tabular} & \begin{tabular}[c]{@{}c@{}}SciDocs\\(EN)\end{tabular} & \begin{tabular}[c]{@{}c@{}}SciFact\\(EN)\end{tabular} & \begin{tabular}[c]{@{}c@{}}TREC\\-COVID\\(EN)\end{tabular} & \begin{tabular}[c]{@{}c@{}}Average\\(EN)\end{tabular} & \begin{tabular}[c]{@{}c@{}}Average\\(EN + ES)\end{tabular} \\ \hline
MrBERT-biomed                        & 77.93                                                                  & \textbf{70.78}                                           & {\ul 89.92}                                                & 51.01                                                    & 9.76                                                 & \textbf{10.05}                                         & {\ul 30.25}                                            & \textbf{48.76}                                            & \textbf{24.71}                                         & \textbf{48.56}                                              \\
\begin{tabular}[l]{@{}l@{}}MrBERT-biomed\\AttMAT (25\%)\end{tabular}            & 77.68                                                                  & 68.24                                                    & 88.67                                                      & 49.66                                                    & 9.66                                                 & 8.87                                                   & 27.81                                                  & 44.08                                                     & 22.60                                                  & 28.02                                                       \\
\begin{tabular}[l]{@{}l@{}}MrBERT-biomed\\AttMAT (50\%)\end{tabular}           & 77.79                                                                  & 68.67                                                    & \textbf{90.05}                                             & \textbf{53.69}                                           & {\ul 9.96}                                           & 9.71                                                   & 28.72                                                  & {\ul 45.98}                                               & {\ul 23.59}                                            & {\ul 29.61}                                                 \\
\begin{tabular}[l]{@{}l@{}}MrBERT-biomed\\AttMAT (75\%)\end{tabular}         & \textbf{78.28}                                                         & {\ul 69.94}                                              & 89.04                                                      & 52.73                                                    & \textbf{10.18}                                       & 9.71                                                   & 30.03                                                  & 39.37                                                     & 22.32                                                  & 28.40                                                       \\
\begin{tabular}[l]{@{}l@{}}MrBERT-biomed\\AttMAT (100\%)\end{tabular}         & {\ul 78.14}                                                            & 69.73                                                    & 89.09                                                      & {\ul 53.19}                                              & 9.58                                                 & {\ul 9.96}                                             & \textbf{30.54}                                         & 41.64                                                     & 22.93                                                  & 28.98                \\ \hline                                      
\end{tabular}
\caption{Evaluation results on different biomedical benchmarks over attention matryoshka in MrBERT-biomed.}
\end{table}

\begin{table}[h!]
\begin{tabular}{lccccccccc}
\hline
\textbf{Model Variant} & \textbf{\begin{tabular}[c]{@{}c@{}}LexBOE\\(ES)\end{tabular}} & \textbf{\begin{tabular}[c]{@{}c@{}}small\\-spanish\\-legal\\-dataset\\(ES)\end{tabular}} & \textbf{\begin{tabular}[c]{@{}c@{}}EURLEX\\(EN)\end{tabular}} & \textbf{\begin{tabular}[c]{@{}c@{}}AILA\\Statutes\\(EN)\end{tabular}} & \textbf{\begin{tabular}[c]{@{}c@{}}legal\_\\summarization\\(EN)\end{tabular}} & \textbf{\begin{tabular}[c]{@{}c@{}}Legal\\Bench\\(EN)\end{tabular}} & \textbf{\begin{tabular}[c]{@{}c@{}}Nano\\Touche\\2020\\(EN)\end{tabular}} & \textbf{\begin{tabular}[c]{@{}c@{}}Average\\(EN)\end{tabular}} & \textbf{\begin{tabular}[c]{@{}c@{}}Average\\(EN + ES)\end{tabular}} \\ \hline
MrBERT-legal & {\ul 96.80} & 38.75 & \textbf{97.33} & \textbf{16.33} & \textbf{55.05} & {\ul 58.04} & 44.74 & \textbf{54.30} & \textbf{58.15} \\
\begin{tabular}[l]{@{}l@{}}MrBERT-legal\\AttMAT (25\%)\end{tabular} & \textbf{96.96} & 39.14 & 97.24 & 13.49 & 53.31 & 56.84 & 45.01 & 53.18 & 57.43 \\
\begin{tabular}[l]{@{}l@{}}MrBERT-legal\\AttMAT (50\%)\end{tabular} & 96.73 & 39.84 & {\ul 97.26} & 10.82 & 53.69 & 56.16 & \textbf{46.39} & 52.86 & 57.27 \\
\begin{tabular}[l]{@{}l@{}}MrBERT-legal\\AttMAT (75\%)\end{tabular} & 96.66 & \textbf{40.31} & 97.25 & {\ul 15.82} & {\ul 54.87} & 56.88 & 45.23 & {\ul 54.01} & {\ul 58.15} \\
\begin{tabular}[l]{@{}l@{}}MrBERT-legal\\AttMAT (100\%)\end{tabular} & 96.74 & {\ul 40.07} & 97.24 & 12.92 & 54.48 & \textbf{58.30} & {\ul 45.34} & 53.66 & 57.87 \\ \hline
\end{tabular}
\caption{Evaluation results on different legal benchmarks over attention matryoshka in MrBERT-legal.}
\end{table}

\end{document}